# High-fidelity Modeling of Full-scale Pressurized Water Reactor Flow Fields for Machine Learning Applications

Logan A. Burnett[a,b], Hyungjun Kim[a,c], Hsien-Cheng Chou[d], Arsha Witoelar[d], Robert A. Brewster[b], Benoit Forget[b], Emilio Baglietto[b], Majdi I. Radaideh*[a,e]

[a]*Department of Nuclear Engineering and Radiological Sciences, University of Michigan, Ann Arbor, MI 48109, United States*

[b]*Department of Nuclear Science and Engineering, Massachusetts Institute of Technology, Cambridge, MA 02139, United States*

[c]*Korea Atomic Energy Research Institute, Daejeon, 34057, South Korea*

[d]*Department of Mechanical Engineering, University of Michigan, Ann Arbor, MI 48109, United States*

[e]*Department of Computer Science and Engineering, University of Michigan, Ann Arbor, MI 48109, United States*

**Abstract**

This work presents a high-fidelity computational fluid dynamics (CFD) and data-driven modeling framework for assembly-level flow characterization in a four-loop pressurized water reactor (PWR). A full lower-plenum and core-inlet domain was constructed using publicly available geometry and operating conditions, enabling transient simulations with pump-induced swirl boundary conditions. The results show that cold-leg swirl and lower-plenum transport generate strongly heterogeneous assembly-wise inlet flow distributions, particularly near the lower core region, while axial resistance and mixing progressively homogenize the flow at higher elevations.

These physics-informed datasets were subsequently used to evaluate machine learning (ML) applications for partial field reconstruction and short-term autoregressive prediction. A 3D convolutional-based inpainting model successfully reconstructed missing assembly-level mass flow rates from partial observations, with errors concentrated in the highly turbulent base (bottom) layer and diminishing significantly in upper layers. Comparative analysis across multiple ML models demonstrates that spatially aware architectures, particularly ConvLSTM, significantly outperform sequence-based (LSTM) and operator-learning (DeepONet) approaches by effectively capturing coupled spatio-temporal dynamics. The study also highlights key challenges, including the sensitivity of inlet flow predictions to turbulence and mesh resolution, as well as the absence of full-scale experimental validation data. Despite these limitations, the results remain consistent with expected physical behavior. Overall, this work establishes high-fidelity CFD as a critical foundation for developing data-driven surrogates, sparse sensing strategies, and future multiphysics coupling frameworks, while outlining directions for scalable, interpretable, and multi-fidelity learning approaches in nuclear reactor modeling.

## 1. Introduction

Thermal-hydraulics (TH) and fluid flow analysis play a central role in reactor safety analysis by providing detailed insights into the complex multiphase flow and heat transfer phenomena that govern reactor operation. These methods enable engineers to predict critical safety parameters such as coolant distribution, cladding temperatures, and pressure drops under both normal and accident scenarios [1]. Unlike purely empirical correlations, Computational Fluid Dynamics (CFD) and advanced TH simulations resolve local velocity, temperature, and turbulence fields, capturing important spatial and temporal variations that influence fuel integrity and system performance [2]. This capability is particularly important in assessing safety margins, evaluating transient scenarios, coupling with other physics simulations (e.g., neutronics), and supporting design improvements for new reactor concepts [3, 4].

Beyond predictive capability, CFD and TH modeling provide a complementary role to system codes by resolving local effects that are otherwise approximated in 1D analyses. For example, CFD has been extensively applied in pressurized water reactor (PWR) mixing studies [5], boiling heat transfer characterization [6], and passive safety

*Corresponding Authors: Logan Burnett (nucleai@umich.edu), Majdi I. Radaideh (radaideh@umich.edu)

**Nomenclature**

| | |
|---|---|
| AI | Artificial Intelligence |
| CFD | Computational Fluid Dynamics |
| CNN | Convolutional Neural Network |
| DNS | Direct Numerical Simulation |
| DT | Digital Twin |
| LES | Large Eddy Simulation |
| LR | Learning Rate |
| LSTM | Long Short-Term Memory |
| ML | Machine Learning |
| PINN | Physics-Informed Neural Networks |
| PWR | Pressurized Water Reactor |
| RANS | Reynolds-Averaged Navier–Stokes |
| SAM | System Analysis Module |
| TH | Thermal-Hydraulics |
| URANS | Unsteady Reynolds-Averaged Navier–Stokes |
| VVUQ | Verification, Validation, and Uncertainty Quantification |

system evaluations [7]. Modern reactor safety evaluations increasingly rely on multi-scale coupling between system thermal-hydraulics [8], subchannel codes [9], and high-fidelity CFD to bridge the gap between integral safety assessments and localized safety phenomena. In this way, CFD and flow analysis not only support regulatory decision-making [10] but also serve as essential tools in validating experimental databases and developing reduced-order or surrogate models for uncertainty quantification [11].

With recent advances in mathematical modeling of physical processes and increased computational power for simulating high-dimensional dynamical systems, CFD has become an essential tool for developing digital twins of nuclear reactors for safety analysis [12, 2]. The robustness of CFD simulations has been widely recognized in the literature [13], particularly in single-phase mixing applications where multidimensional flow effects play a significant role [14]. CFD is especially valuable for addressing limitations of system codes, which often fail to capture local transfer rates of momentum, energy, and mass. CFD simulations require rigorous validation, and guidance from Idaho National Laboratory (INL) outlines best practices for quantifying uncertainties associated with discretization schemes, turbulence modeling, and iterative convergence [15**?** ]. The advantages of CFD over lower-dimensional approaches have been demonstrated in several studies. For instance, improved accuracy in capturing multidimensional flow mixing in a T-junction compared to the 1D thermal-hydraulics code TRACE has been shown [16]. Similarly, steady-state CFD modeling of a PWR fuel assembly has been validated against experimental benchmarks, supporting its predictive capability [17]. Verification and validation studies for both single-phase and two-phase CFD simulations in PWR fuel assemblies have also been conducted, including performance assessments of Mixing Vane Spacer Grids (MVG) [18].

CFD simulations of reactor pressure vessels have been used to analyze the impact of internal structural configurations, such as perforated drums, on mass flow distribution at the reactor core [19]. Although Direct Numerical Simulation (DNS) and Large Eddy Simulation (LES) provide high fidelity, their computational expense has motivated the development of Reynolds-Averaged Navier–Stokes (RANS) approaches. In this work, CFD modeling is performed using the `STAR-CCM+` software package [20], with turbulence modeling details provided in Section 3.

Numerous studies have investigated single-phase coolant flow through reactor fuel rod bundles, as in [21]. While CFD offers high-fidelity, numerical methods such as LES and DNS demand more computing resources than conventional system codes, making them difficult to use for nuclear reactor safety analysis [**?** ]. To address this challenge, hybrid modeling strategies have been explored, such as the Structure-based (STRUCT-$\epsilon$) turbulence model, which aims to balance fidelity and computational efficiency [22, 23]. In parallel, a paradigm shift in nuclear reactor simulation has been underway, moving from separate codes to modular, extensible multiphysics frameworks. A leading example is the Multiphysics Object-Oriented Simulation Environment (MOOSE) from Idaho National

Laboratory [24]. Also, the System Analysis Module (SAM) from Argonne National Laboratory, a modern tool for advanced reactor safety analysis, is an example of proper application [25]. Beyond the MOOSE Framework, additional high-fidelity modeling studies of heat pipe microreactors—encompassing both single-physics and coupled multiphysics analyses—have been demonstrated using OpenFOAM [26, 27].

Reinforcement learning has been applied to decouple core power and axial power distribution control in large pressurized water reactors [28], while LSTM- and CNN-LSTM-based frameworks have shown strong performance for fault diagnosis and cross-operating-condition prediction in small modular reactors [29, 30]. In addition, recent hybrid approaches combining data-driven learning with physics-based constraints have improved the robustness and physical consistency of reactor coolant loop modeling [31]. These studies highlight the increasing integration of AI, physics-informed learning, and digital twin concepts in advanced reactor thermal-hydraulic applications.

Machine learning (ML) has emerged as a promising approach to reduce the computational cost of high-fidelity thermal-fluid and multiphysics simulations while maintaining predictive accuracy. For instance, Convolutional Neural Networks (CNNs), trained on high-fidelity DNS data, can predict complex turbulent flow fields with less resources than CFD [32, 33, 34]. This shows the potential to reduce the computational cost of high-fidelity simulations by LES or DNS. Furthermore, recent advancements include the use of Dense-CNN/LSTM (Long Short-Term Memory) architecture to predict turbulent viscosity during a reactor transient [33]. Another approach involves Physics-Informed Neural Networks (PINNs), which embed the governing differential equations directly into the model's training process. A recent study demonstrated that a Transfer Learning-enhanced PINN (TL-PINN) could predict nuclear reactor transients with up to a two orders of magnitude acceleration in training time compared to a standard PINN [35]. In nuclear engineering applications, ML-driven surrogate models have demonstrated strong capability in both steady-state and transient analyses. For example, neural-based time-series forecasting methods have been used to predict loss-of-coolant accident (LOCA) dynamics, enabling rapid approximation of system behavior without repeatedly solving full-order models [36]. Similarly, deep learning methods [37, 38] as well as Gaussian process–based surrogate models [39] provide uncertainty-aware emulation of complex simulations to enable core design [40], sensitivity analysis and uncertainty propagation to model the onset of natural circulation in high-temperature gas-cooled reactors (HTGRs) [41], and reactivity control and criticality search for microreactors [42, 43].

Recent advances extend these approaches to multifidelity frameworks, where surrogate models integrate low- and high-fidelity data to efficiently capture complex thermal-fluid phenomena, such as depressurized loss of forced cooling (DLOFC) scenarios in HTGRs [44]. Complementary efforts in generative modeling, including data-efficient generative adversarial networks (GANs), have shown promise in augmenting limited datasets for critical heat flux predictions [45], while variational digital twin frameworks introduce probabilistic representations for improved system monitoring and prediction for Depressurized Conduction Cooldown experiments [46]. Recurrent nerual networks which will be used in this study showed a strong performance to drive digital twin development of integrated thermal energy systems [47]. Recent studies further demonstrate the growing role of machine learning in nuclear reactor monitoring and control.

This research focuses on developing an open-source pipeline for transient, high-fidelity modeling of inlet flow conditions in pressurized water reactors (PWRs). The framework is designed to support data-driven applications, particularly those leveraging machine learning. The goal is to enable multi-fidelity modeling of inlet flow conditions, where transient simulations generate inlet flow fields that can serve as training data for data-driven models. These models can then be applied to tasks such as spatiotemporal analysis, multi-fidelity surrogate modeling, and multiphysics coupling. Accordingly, the key contributions of this study can be summarized as:

1. The team developed a STAR-CCM+ model of the lower plenum for a full-scale commercial PWR unit,

previously represented in the BEAVRS benchmark [48]. Core dimensions and lower plenum configurations were compiled from a wide range of public sources, including information on fuel assembly modeling from BEAVRS.

2. A representative pump swirl pattern was implemented at the cold legs to better represent heterogeneous inlet conditions. This approach captures flow asymmetries that may drive power tilts by inducing non-uniform flow distributions and variable hydraulic forces at the fuel assembly inlets.
3. Transient CFD models were constructed with varying mesh refinements to generate datasets characterizing assembly-level flow conditions in both space and time. These datasets were then used to train spatiotemporal machine learning models—including convolutional neural networks, long short-term memory (LSTM), convolutional LSTM (ConvLSTM), and deep neural operators (DeepOnet)—which demonstrated excellent predictive accuracy and significantly reduced computational costs in forward simulations.
4. The team plans to release both the raw CFD models and the generated datasets as open-access resources. This will enable broader verification & validation, foster the creation of more complex datasets, and promote the efficient use of high-fidelity simulation data to train data-driven models for multiphysics coupling applications.

The remainder of this paper is structured as follows. Section 2 introduces the design data collected to model the lower plenum of a commercial PWR and explains how these data were implemented into a steady-state CFD model using STAR-CCM+. Section 3 describes the conversion of the steady-state model into a transient simulation and the incorporation of pump swirl effects at the cold legs. Section 4 presents and discusses the CFD simulation results, including transient CFD simulations, while section 5 presents two ML applications on the CFD datasets generated in this study and the limitations and research implications of this work. Finally, Section 6 summarizes the key findings and highlights potential directions for future research.

## 2. Reactor Modeling and Simulation

### *2.1. Design Data*

We constructed a four-loop PWR flow-domain CAD in SolidWorks using publicly available references from BEAVRS geometric data for vessel/barrel/baffle/core envelope [48], an ORNL system-definition document for operating conditions and core internals dimensions [49], and the NRC Westinghouse Technology Systems Manual for selected plate and support features [50]. The modeled fluid domain extends from the four cold-leg nozzles, through the downcomer and lower plenum, up to the top-of-fuel plane to match the CFD inlet to the core (fuel-assembly) region. Table 1 summarizes the main geometrical specifications of the core used in building the CFD model. A visualization of the 4-loop PWR geometry developed from the publicly available data in this study is illustrated in Figure 1.

The workflow included (1) extracting dimensions and topology from the sources; (2) building a parameterized sketch (vessel/barrel radii, baffle gap, core envelope); (3) generating revolved and extruded features (vessel, barrel, baffle, lower plenum) for the four-loop layout; (4) representing core support components and instrumentation/guide tubes with values extracted from design schematics; (5) defining fluid–solid interfaces and trimming to the flow domain. We use dimensions directly where references provided ranges or nominal values and extract other dimensions from images in the documents.

STAR-CCM+ is used to model the core in this analysis. A mesh was created using the 3D-CAD model of the plant. The original simulation solves for water with constant density and a segregated fluid isothermal configuration. It utilizes a Reynolds-averaged Navier-Stokes turbulence model and a realizable k-$\epsilon$ two-layer model. Geometry and mesh quantities were parameterized to other global quantities. This automated most of the work needed to change the model parameters, allowing for ease in the mesh sensitivity studies. In this model, the Base Mesh Size (see Table 2) acts as the base parameter upon which the other parameters are dependent.

Table 1: Design and system parameters with values in both Imperial and SI units.

| Parameter | Imperial | SI | Source |
| --- | --- | --- | --- |
| Inlet Coolant Temperature (°F/°C) | 557.6 °F | 292 °C | [49] |
| System Pressure (psia/Pa) | 2250 psia | $1.55 \times 10^7$ Pa | [48] |
| Support Plate Thickness (in/cm) | 21.2 in | 53.85 cm | [50] |
| Lower Core Plate Thickness (in/cm) | 2.46 in | 6.26 cm | [50] |
| Tie Plate Thickness (in/cm) | 3.57 in | 9.07 cm | [50] |
| Secondary Core Support Thickness (in/cm) | 4.29 in | 10.9 cm | [50] |
| Fuel Assembly Length – Lower Nozzle to Top of Fuel (in/cm) | 159.8 in | 405.9 cm | [49] |
| Fuel Assembly Pitch (in/cm) | 8.644 in | 21.5 cm | [49] |
| Rod Pitch (in/cm) | 0.496 in | 1.26 cm | [49] |
| Instrument Thimble Upper OD (in/cm) | 0.482 in | 1.22 cm | [49] |
| Instrument Thimble Lower OD (in/cm) | 0.429 in | 1.09 cm | [49] |
| Guide Tube OD (in/cm) | 0.482 in | 1.22 cm | [49] |
| Baffle Width (in/cm) | 0.875 in | 2.2225 cm | [48] |
| Number of Cold Legs | — | 4 (dimensionless) | [49] |
| Core Barrel Inner Radius (in/cm) | 74.0 in | 187.96 cm | [48] |
| Core Barrel Outer Radius (in/cm) | 76.25 in | 193.675 cm | [48] |
| Pressure Vessel Inner Radius (in/cm) | 86.5 in | 219.71 cm | [48] |
| Pressure Vessel Outer Radius (in/cm) | 95.0 in | 241.3 cm | [48] |
| Total Mass Flow Rate (lbm/s / kg/s) | 39231.3 lbm/s | 17790 kg/s | - |

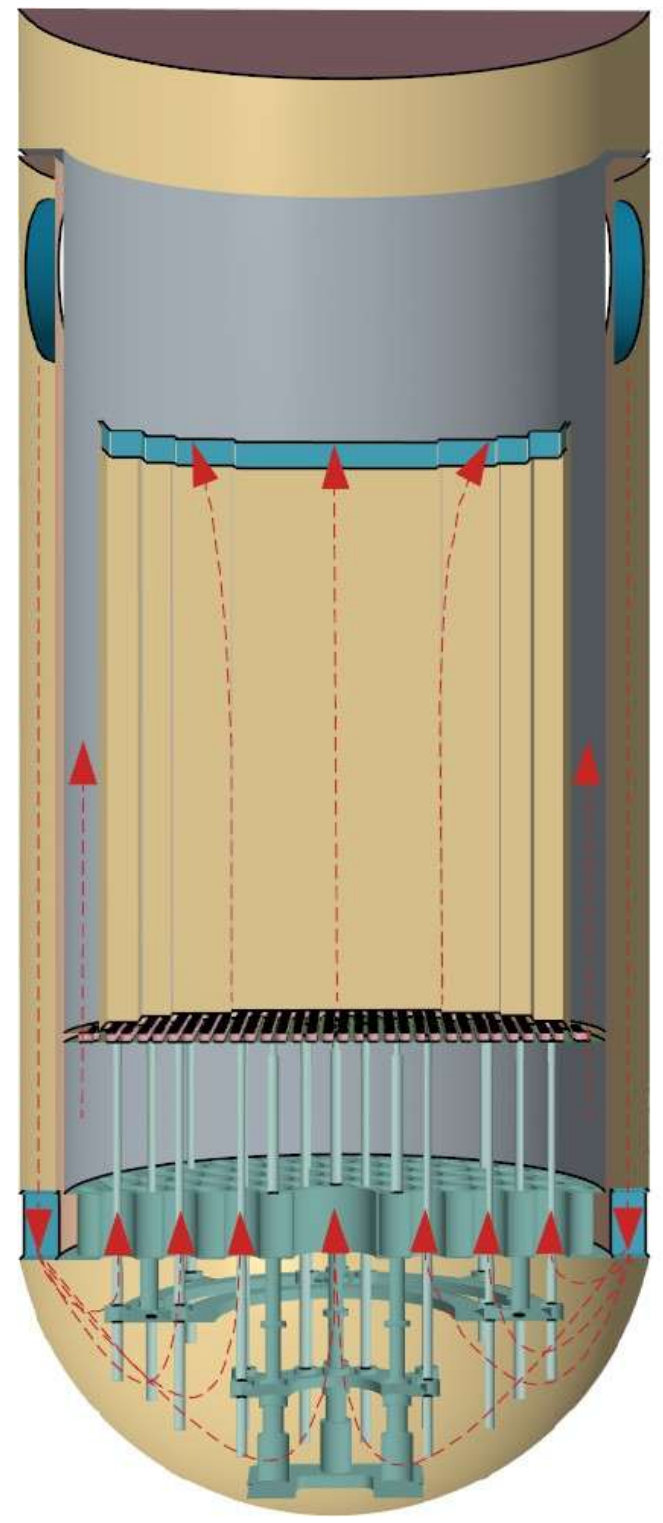
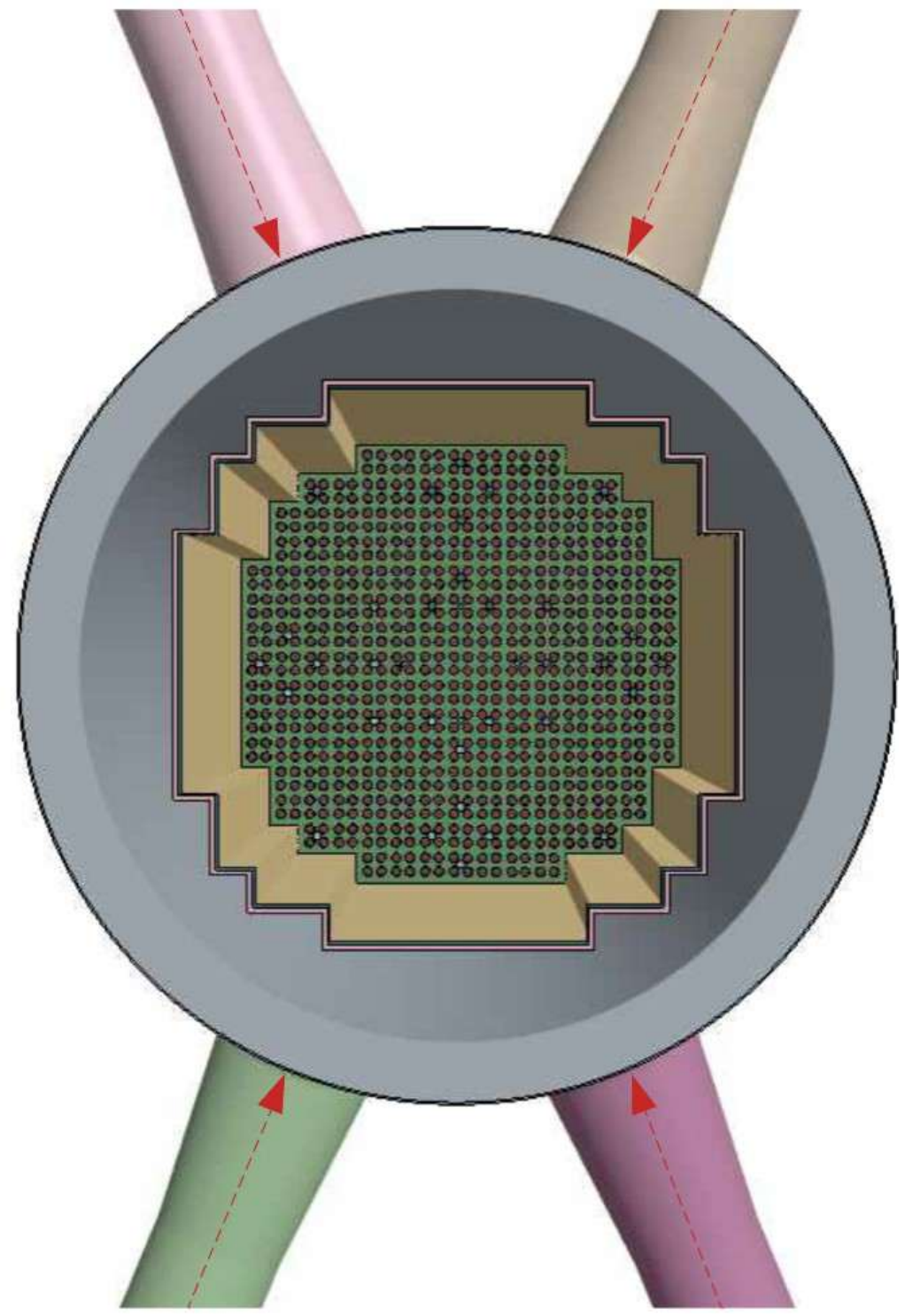

Figure 1: Visualization of 4-loop PWR geometry developed from publicly available data.

Due to the complexity of the fuel assembly geometry, each fuel assembly was recreated as a porous media representation using the Forchheimer equation,

$$\frac{dP}{dx_i} = (\alpha_i|v_i| + \beta_i)\, v_i, \tag{1}$$

where $P$ is the pressure, $x_i$ is the spatial coordinate in direction $i$, and $v_i$ is the superficial velocity component through the porous medium in that direction. The coefficients $\alpha_i$ and $\beta_i$ represent the inertial and viscous resistance terms, respectively, and were obtained from curve fits of pressure-versus-flow-rate test data for the fuel assembly. Figure 2 presents the porous media representation used in the STAR-CCM+ model alongside a diagram of the original fuel assembly geometry used in the PWR unit of interest [49]. Table 2 lists the corresponding parameters used for the porous media fuel assemblies. Sections of the mesh are shown in Figure 3, including the porous media representation for the fuel assemblies and the volumetric mesh for the reactor vessel.

Table 2: Fuel assembly porous media parameters (Imperial and SI units).

| **Parameter** | **Imperial** | **SI** | **Definition** |
|---|---|---|---|
| Axial Porous Inertial Resistance | 410.0 lbm/ft$^4$ | 5949.0 kg/m$^4$ | Resistance coefficient for axial inertial losses in porous medium |
| Axial Porous Viscous Resistance | 169.3 lbm/(ft$^3$·s) | 2 428.0 kg/(m$^3$·s) | Resistance coefficient for axial viscous drag |
| Lateral Porous Inertial Resistance | 2071.9 lbm/ft$^4$ | 30061.4 kg/m$^4$ | Resistance coefficient for lateral inertial losses |
| Lateral Porous Viscous Resistance | 0.0 lbm/(ft$^3$·s) | 0.0 kg/(m$^3$·s) | Resistance coefficient for lateral viscous drag |

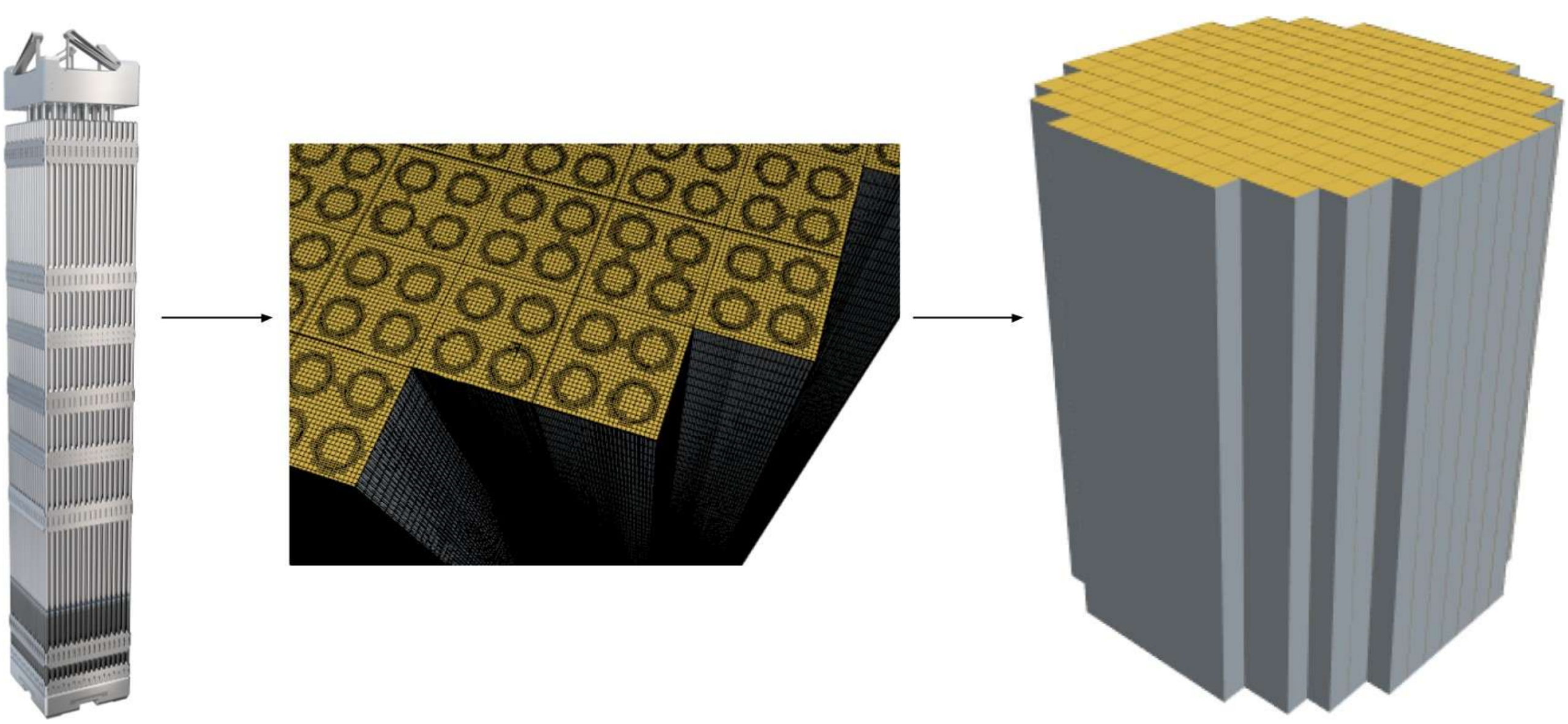

Figure 2: PWR fuel assembly geometry to porous media representation used in the CFD model

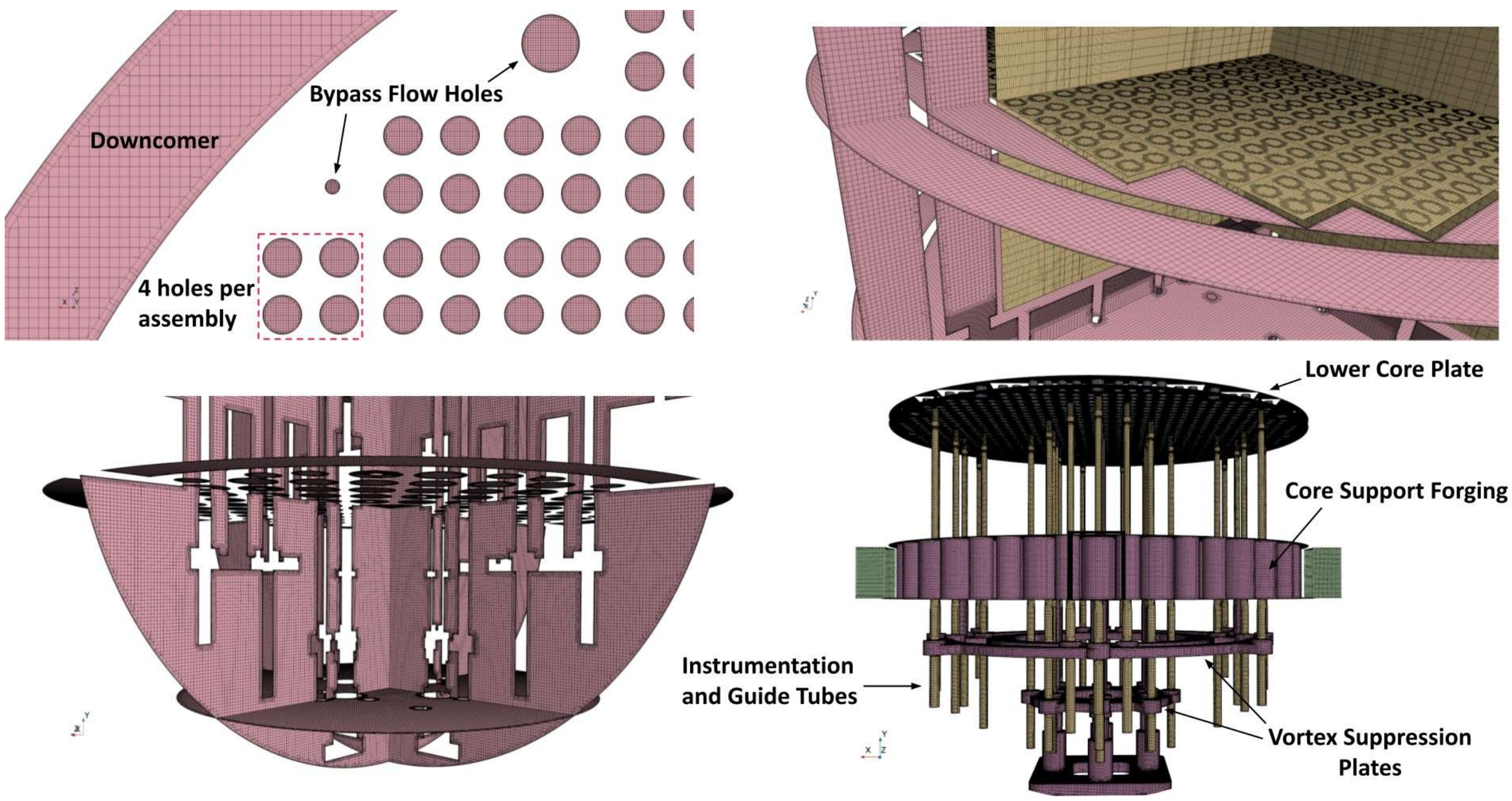

Figure 3: Mesh sections of reactor internals including the lower plenum, downcomer, and fuel assemblies.

## 3. Methodology

### *3.1. Turbulence Modeling*

Classical unsteady Reynolds-averaged Navier–Stokes (URANS) closures retain the time derivative in the turbulence transport equations but still rely on the eddy-viscosity hypothesis and an assumption of equilibrium between production and dissipation of turbulent kinetic energy. In complex reactor flows with jets, separation or swirl, this assumption breaks down. URANS models tend to over-predict turbulent viscosity, leading to unphysically large eddies and overly diffused temperature and momentum fields [51]. These deficiencies motivate the development of second-generation URANS (2G-URANS) models that preserve URANS robustness while locally resolving unsteady structures.

The baseline closure underpinning STRUCT-$\varepsilon$ is an anisotropic $k$–$\varepsilon$ model with a non-constant $C_\mu$ coefficient that avoids negative stresses and augments the linear stress–strain relationship with quadratic and cubic terms [52]. In this framework, the turbulent viscosity is $\nu_\tau = C_\mu k^2/\varepsilon$, and the transport equations for $k$ and $\varepsilon$ are

$$\frac{\partial k}{\partial t} + u_j \frac{\partial k}{\partial x_j} = \frac{\partial}{\partial x_j} [(\nu + \frac{\nu_\tau}{\sigma_k}) \frac{\partial k}{\partial x_j}] + P_k - \varepsilon, \quad (2)$$

$$\frac{\partial \varepsilon}{\partial t} + u_j \frac{\partial \varepsilon}{\partial x_j} = \frac{\partial}{\partial x_j} [(\nu + \frac{\nu_\tau}{\sigma_\varepsilon}) \frac{\partial \varepsilon}{\partial x_j}] + C_{\varepsilon 1} \frac{\varepsilon}{k} P_k - C_{\varepsilon 2} \frac{\varepsilon^2}{k}. \quad (3)$$

Here $k$ denotes the turbulent kinetic energy, $\varepsilon$ denotes the turbulent dissipation rate, $x_j$ is the $j$-th spatial coordinate, $u_j$ is the $j$-th component of the Reynolds-averaged velocity field, $\nu$ is the molecular kinematic viscosity, $\nu_\tau$ is the turbulent kinematic viscosity, $C_\mu$ is the turbulent-viscosity coefficient, $P_k$ denotes the production of turbulent kinetic energy, and $\sigma_k$, $\sigma_\varepsilon$, $C_{\varepsilon 1}$, $C_{\varepsilon 2}$ are standard model constants. This anisotropic URANS closure improves predictions of swirling and curved flows but still assumes scale separation and isotropic dissipation [53, 54].

STRUCT models introduce a structure-based resolution criterion. A resolved time scale $\tau_r$ is defined from the second invariant of the resolved velocity-gradient tensor,

$$Q = -\tfrac{1}{2}\,\bar{A}_{ij}\bar{A}_{ji}, \qquad \bar{A}_{ij} = \frac{\partial \bar{u}_i}{\partial x_j},$$

and expressed as $\tau_r^{-1} = \sqrt{|Q|}$ [52]. The modeled time scale is $\tau_m = k/\varepsilon$; their ratio $\alpha = \tau_r/\tau_m$ indicates where resolved and modeled scales overlap. In regions of large $\alpha$, the eddy viscosity is reduced by a function $f_m(\alpha)$, which transitions the model from a URANS-like state to a scale-resolving regime without reference to the mesh.

STRUCT-$\varepsilon$ applies the structure-based idea by modifying the dissipation equation. The $\varepsilon$ transport equation becomes

$$\frac{\partial \varepsilon}{\partial t} + u_j \frac{\partial \varepsilon}{\partial x_j} = \frac{\partial}{\partial x_j}\left[\left(\nu + \frac{\nu_\tau}{\sigma_k}\right)\frac{\partial \varepsilon}{\partial x_j}\right] + C_{\varepsilon 1}\frac{\varepsilon}{k}P_k - C_{\varepsilon 2}\frac{\varepsilon^2}{k} + C_{\varepsilon 3}\,k\,|\Pi|, \tag{4}$$

where $|\Pi|$ is the magnitude of the second principal invariant of the resolved velocity-gradient tensor. The additional term $C_{\varepsilon 3}\,k\,|\Pi|$ is absent in standard URANS models. It suppresses dissipation wherever the resolved strain or rotation is large, thereby reducing $\nu_\tau$ and forcing the simulation to resolve eddies of roughly the integral length scale [51]. This simple modification allows STRUCT-$\varepsilon$ to deliver finer momentum structures and more accurate temperature predictions in stratified sodium experiments and thermal striping in T-junctions while retaining the computational cost of URANS [51, 55].

### 3.2. Swirl Modeling

The coolant inlet flow to the Pressurized Water Reactor (PWR) comes from the outlet of Reactor Coolant Pumps (RCPs), which are known to introduce non-uniform inlet conditions at the inlet of cold legs [14], consequently leading to power tilts or non-uniform flow rate distributions at the axial core inlet. RCPs are centrifugal pumps [14] that induce swirl patterns at their outlets. Hence, to artificially reproduce these effects, we implement a swirling boundary condition at the inlet of all four cold legs. In this study, we implement this via defining a user-defined function (UDF), which requires a functional form for the boundary condition to be implemented, and is presented below in the polar coordinate form. The parameter $\alpha_s$ in the equation represents the tangential to axial velocity ratio or the swirl strength parameter, this parameter defines the extent of the swirl strength at the inlet, meaning, ($\alpha_s = 0$) would effectively produce no swirl at the boundary. This parameter is a variable quantity and can act as an inference parameter with respect to its corresponding effects at the axial core inlet.

$$\mathbf{u}(r,\theta) = \alpha_s u_{\text{axial}}\mathbf{e}_\theta + u_{\text{axial}}\mathbf{e}_z, \qquad u_\theta(r) = \alpha_s u_{\text{axial}}. \tag{5}$$

$$\mathbf{e}_r = (\cos\theta, \sin\theta, 0), \qquad \mathbf{e}_\theta = (-\sin\theta, \cos\theta, 0), \qquad \mathbf{e}_z = (0, 0, 1). \tag{6}$$

Here, $\mathbf{u}(r,\theta)$ is the imposed inlet velocity vector expressed in local polar coordinates, $r$ is the radial coordinate measured from the center of the cold-leg inlet, $\theta$ is the azimuthal coordinate in the inlet cross section, $\alpha_s$ is the swirl strength parameter defined as the tangential-to-axial velocity ratio, $u_{\text{axial}}$ is the prescribed axial inlet velocity, $u_\theta(r)$ is the tangential velocity component, and $\mathbf{e}_r$, $\mathbf{e}_\theta$, and $\mathbf{e}_z$ are the local radial, tangential, and axial unit vectors, respectively.

The polar coordinate functional form translates to the following Cartesian form, which is used to define the field function in `STAR-CCM+` [20]. An additional coefficient, $\epsilon$, is included in the denominator to avoid singular behavior at the center of the circular inlet cross section. Figure 4 demonstrates a side view of the cold leg and the swirl shape applied as a boundary condition. The coordinate system is defined such that the $z$-axis is normal to the inlet cross-sectional plane and the $x$-$y$ axes lie within the inlet cross section. The equivalent Cartesian form used in the solver field function is

$$[U_x, U_y, U_z] = \left[ -\alpha_s u_{axial} \frac{y}{\sqrt{x^2 + y^2 + \epsilon}} \ , \ \alpha_s u_{axial} \frac{x}{\sqrt{x^2 + y^2 + \epsilon}} \ , \ u_{axial} \right] \tag{7}$$

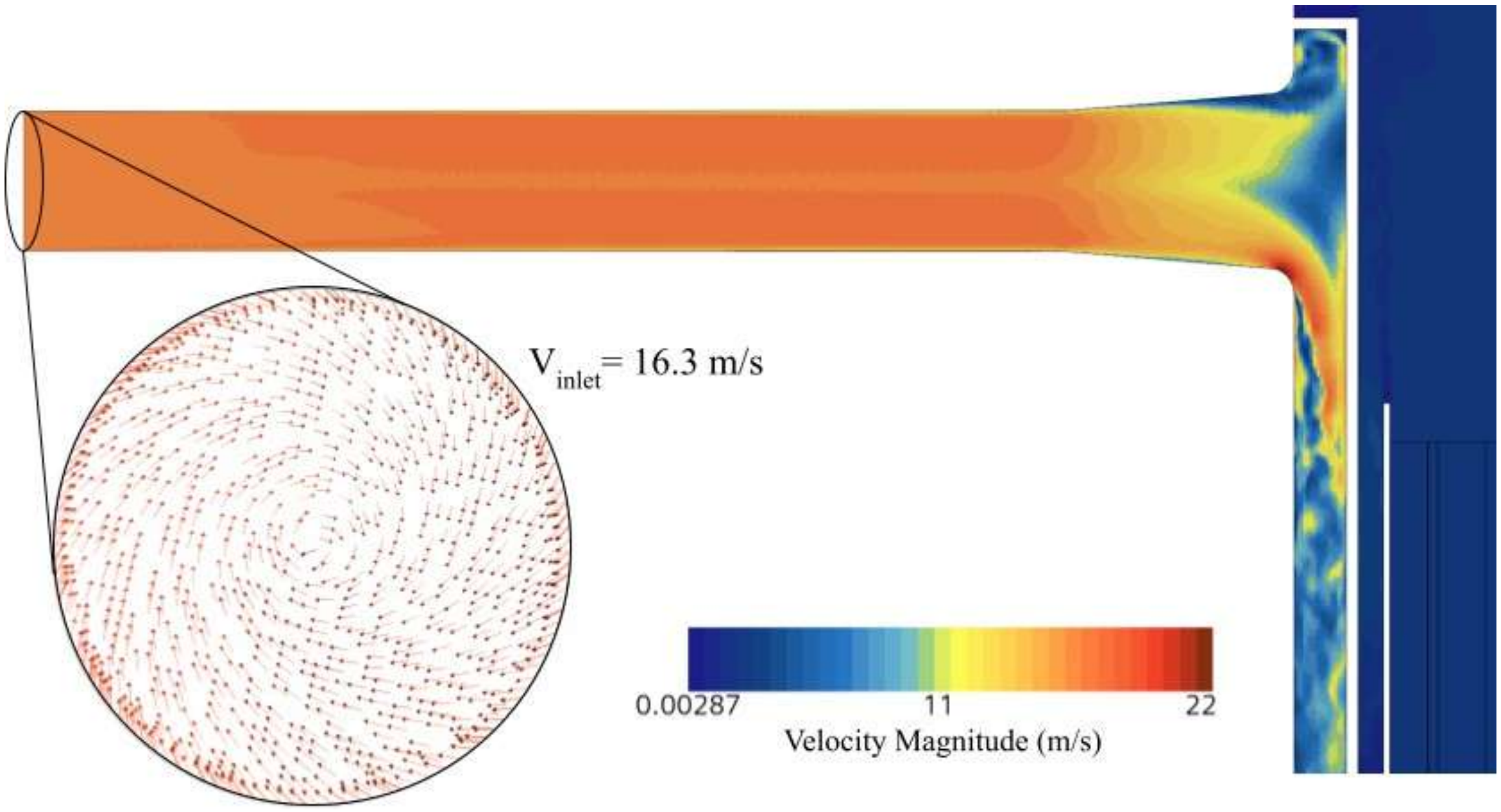

Figure 4: Cold leg inlet cross-sectional view (side view)

### 3.3. Transient modeling

The lower-plenum flow field is inherently unsteady due to the interaction of the imposed cold-leg swirl, geometric redirection, recirculation, and turbulent mixing within the reactor vessel. A steady-state simulation may capture the mean flow structure, but it cannot represent the temporal redistribution of assembly-wise mass flow rates associated with these unsteady turbulent dynamics. Therefore, transient CFD simulations were conducted to quantify the time-dependent mass flow rate distribution across the fuel assemblies in the core.

The transient flow simulation is then analyzed as follows. A single sensor plane is created and constrained by each fuel assembly at the fluid entrance for said fuel assembly. This process is then repeated at 0.5 meter increments in the axial direction of the PWR until 9 total layers of sensor planes are generated. This provides a total of 193 sensor planes per layer, 1737 sensor planes in total. The sensor planes are programmed to record average mass flow rates within each time step. The model first undergoes CFD simulation until a physical time equivalent of 10 seconds with a time step of 0.0005 seconds, each timestep consisting of 5 internal iterations. This process ensures full development of the fluid profile before data is taken. We then record data using the same time step size between 10-15 seconds, recording the mass flow rate in all 193 assemblies. All parameters used in the transient simulation are given in Table 3.

Table 3: Transient Simulation Parameters

| Parameter | Value | Definition / Notes |
|---|---|---|
| Total Simulation Time | 15 s | Total transient simulation duration |
| Initial Flow-Development Period | 0–10 s | Period used for flow-field development before data collection |
| Data Collection Interval | 10–15 s | Time window used to record assembly-wise mass flow rates |
| Time Step Size ($\Delta t$) | 0.0005 s | Unsteady CFD time step |
| Internal Iterations per Step | 5 | Solver iterations per physical time step |
| Planar Sensor Count Per Layer | 193 planes | One area-averaged sensor plane per fuel assembly |
| Planar Sensor Layer Count | 9 layers | Axial sensor layers spaced by 0.5m |
| Total Sensor Planes | 9 × 193 = 1737 planes | Total assembly-level sensor planes monitored |
| Solver Type | SIMPLE-based segregated solver | Collocated finite-volume solver with Rhie–Chow interpolation |
| Temporal Discretization | Second-order implicit | Three-time-level backward differencing scheme |
| Density Model | Constant density | Single-phase incompressible |
| Thermal Treatment | Isothermal | Energy equation and buoyancy effects not modeled |
| Turbulence Model | STRUCT-$\varepsilon$ | Structure-based URANS closure |
| Wall Treatment | High $y^+$ wall treatment | Wall-function treatment for the near-wall mesh |
| Dimensionality | Three-dimensional | Full 3D lower-plenum and core-inlet flow domain |
| Output Frequency | Every time step during data collection | Mass flow rates recorded over the 10–15 s interval |
| Data Averaging | Area-averaged mass flow rate | Used for time histories, heatmaps, and ML datasets |

### *3.4. Mesh Sensitivity study*

Due to the complexity and high computational cost of the transient model developed in this study, the authors adopted a simplified approach by testing simulations using different mesh resolutions while leveraging extensive prior CFD expertise. After our preliminary mesh analysis, three primary models were evaluated and executed according to the procedure outlined in the previous sections. These models are summarized in Table 4, each assigned a unique name that will be referenced throughout this work. The flow results of the lower-fidelity models are compared against those of the high-fidelity (reference) model to assess the impact of mesh coarsening on the accuracy of the flow predictions. Note that subsequent analyses in this study are carried out using the highest-fidelity model, under the assumption that it provides the most accurate representation.

### *3.5. Computing Resources*

This study leveraged an internal server at the University of Michigan to perform CFD transient simulations across the selected mesh resolutions. The highest-fidelity case was executed using 200 cores and required approximately 20 days to complete, whereas the lowest-fidelity simulation finished in about 6 days. The server is powered by two AMD EPYC 9654 processors, each supporting 192 core threads at clock speeds ranging from 2.4 to 3.7 GHz, yielding a total of 384 core threads, of which 200 were utilized in this study. It is also equipped with four NVIDIA RTX 6000 Ada Generation GPUs and 1536 GB of DDR5 memory, which were employed for the machine learning tasks discussed in Section 5.

Table 4: Description of the different CFD Models used in the mesh sensitivity study

| Model Name | Base Mesh Size |
|---|---|
| High-fidelity or fine mesh ($FA_{pitch}$ 14) | 0.6174 in / 1.5683 cm |
| Medium-fidelity or medium mesh ($FA_{pitch}$ /12) | 0.7203 in / 1.8296 cm |
| Low-fidelity or coarse mesh ($FA_{pitch}$ 10) | 0.8644 in / 2.1956 cm |

## 4. CFD Simulation Results

In this section, we present the transient CFD results. *Unless stated otherwise, the high-fidelity model mentioned in Table 4 is presented in this section*. The CFD results in Figure 5 demonstrate that the imposed cold-leg swirl and lower-plenum geometry produce a strongly three-dimensional inlet flow field rather than a spatially uniform core inlet condition. The side-view velocity magnitude field shows that the largest coolant velocities occur in the downcomer and lower-plenum regions, where the incoming cold-leg momentum is redirected toward the core support region. This behavior is expected because the coolant must turn from the annular downcomer into the lower plenum before entering the fuel assembly region. The lower-plenum flow is therefore characterized by localized high-velocity structures, recirculation, and mixing before the flow enters the porous-media representation of the fuel assemblies.

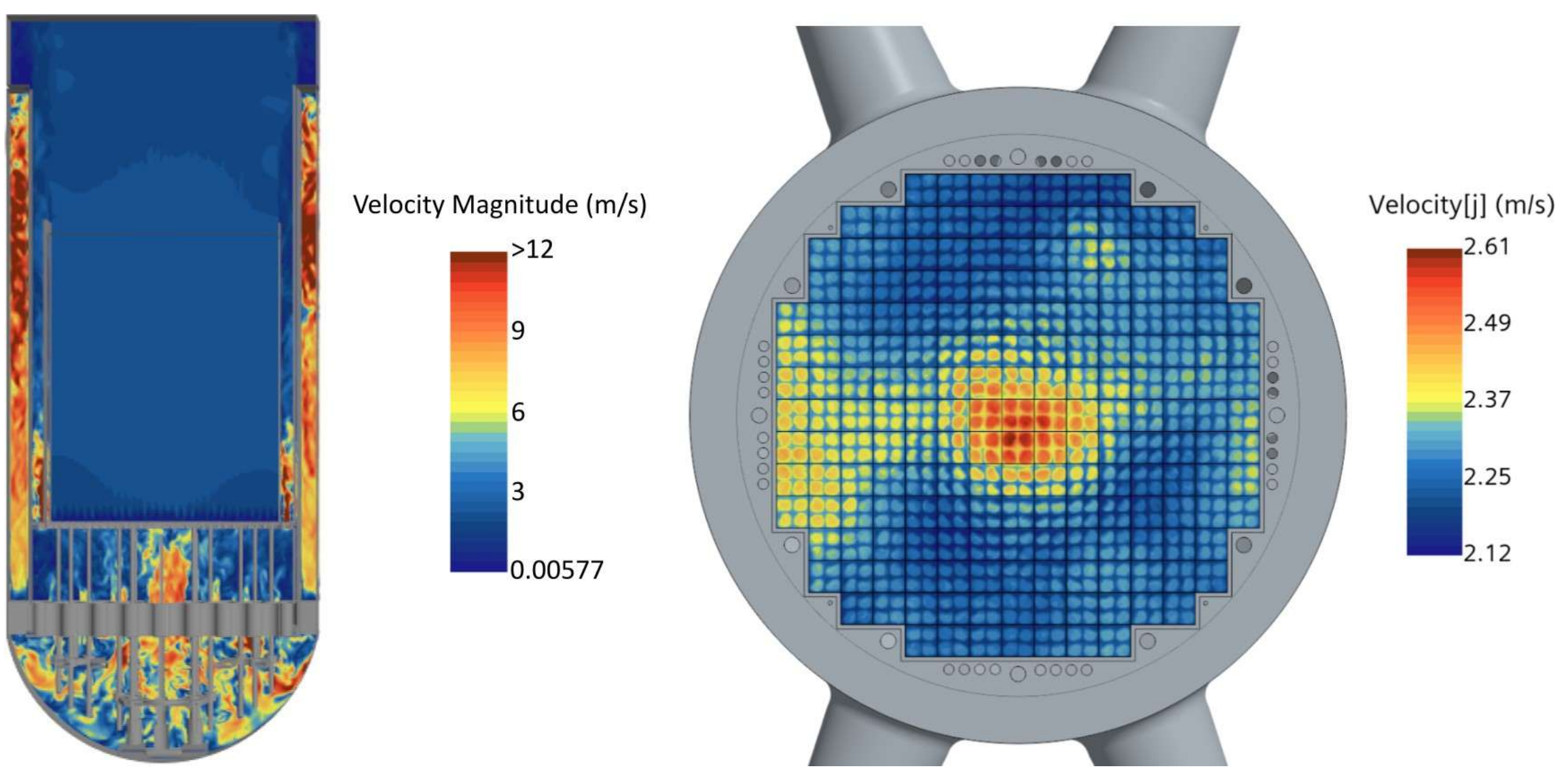

Figure 5: Radial cross section of the reactor vessel colored by velocity magnitude (left) and axial cross section of the core just above the core inlet (right). Both are snapshots of the flow fields at the end of the transient simulation (15s) for the **high-fidelity** model.

The axial core-inlet velocity field in Figure 5 further shows that these lower-plenum effects persist at the assembly inlet plane. The vertical velocity component is not distributed uniformly across the core. Instead, the field contains a pronounced central high-velocity region and additional asymmetric structures distributed across the assembly map. These spatial patterns indicate that the inlet condition to the core retains memory of the cold-leg swirl and lower-plenum redirection, even after partial mixing in the vessel internals. This result is important because many reduced-order or system-level analyses assume comparatively smooth or prescribed inlet flow distributions. The present CFD model shows that the actual inlet condition can contain assembly-scale heterogeneity that may affect local coolant delivery, hydraulic forcing, and downstream thermal-hydraulic behavior.

The axial evolution of the assembly-level mass flow rate is shown in Figure 6. Near the lower entrance of the core, the mass flow distribution exhibits substantial spatial variation across the fuel assemblies. The base plane contains strong localized peaks and depressions, consistent with the non-uniform axial velocity field observed at the core inlet. As the coolant moves upward through the fuel assembly region, the spatial gradients decrease sharply. By the upper planes, the mass flow field becomes significantly smoother and the range of assembly-to-assembly variation is much smaller. This axial damping is consistent with the role of the porous fuel assembly region. Axial resistance and continued mixing reduce the magnitude of inlet maldistribution as the flow progresses through the core.

It is important to note that each axial layer in Figure 6 is plotted with an individual colorbar. This choice is necessary because the range of mass flow values decreases substantially with increasing axial distance from the core entrance. Therefore, the figure should not be interpreted as showing the same absolute variation at every elevation. Instead, one can observe change in spatial structure as a function of height, or distance from the core inlet. The lower planes contain stronger localized heterogeneity, while the upper planes retain weaker, smoother remnants of the original inlet pattern. This distinction is important for downstream data-driven modeling because a model trained on these fields must learn both the high-gradient inlet behavior and the progressively homogenized flow fields at higher elevations.

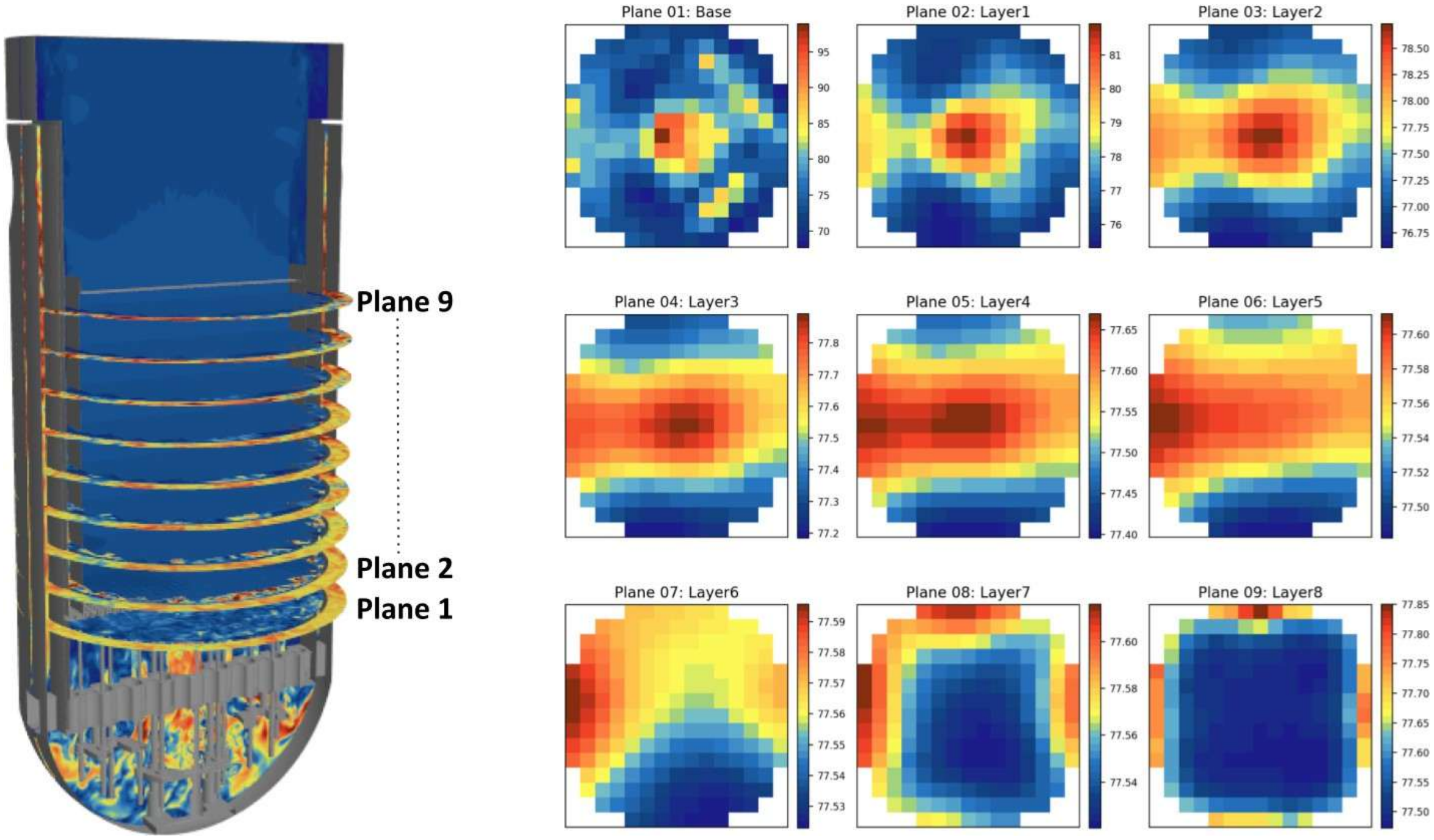

Figure 6: Selected axial planes for monitoring coolant mass flow rate distribution (kg/s) across the fuel assemblies from the lower entrance of the core near Plane 1 to the upper exit of the core near Plane 9. Each axial layer has an individual colorbar given that the range of mass flow rate values decreases sharply with increasing axial distance from the core entrance. The flow fields shown in this figure correspond to the **high-fidelity** model.

The axial evolution of the assembly-level mass flow rate in Figure 6 shows that the inlet distribution is strongest near the lower entrance of the core and decreases rapidly with axial distance. To assess whether these observed structures are robust to mesh resolution, we compare the base-layer mass flow distributions obtained from the

high-, medium-, and low-fidelity meshes. The comparison is performed using both the maximum percent difference over the transient and the time-averaged percent difference at each fuel assembly location.

Figure 7 shows that mesh-resolution effects are spatially localized rather than uniformly distributed across the core inlet. Two types of plots are provided in Figure 7:

- Maximum Difference plots: These plots compare the flow rates between two fidelity levels for each assembly across all time steps, and report the **maximum observed difference** over the entire transient.
- Time-Averaged Error plots: These plots compare the flow rates between two fidelity levels for each assembly across all time steps, and report the **mean difference** computed over the entire transient.

The largest local deviations occur in the maximum-difference fields, where isolated assemblies exhibit errors on the order of 21–27%, depending on the mesh comparison. These localized extrema are most pronounced when the fine mesh is compared against the coarse mesh, which is expected because this comparison represents the largest change in spatial resolution. However, the corresponding core-average errors remain much smaller, with average values near the 6% level. This indicates that the dominant flow structures are broadly preserved across the mesh resolutions, while local assembly-level differences persist in regions with stronger gradients or more complex inlet structure.

The time-averaged difference maps shown in Figure 7 provide a more representative measure of the persistent mesh-induced differences throughout the transient. In comparison to the maximum-difference maps, the time-averaged fields appear smoother and generally exhibit smaller local deviations. All three fidelity comparisons yield relatively small core-averaged errors of approximately 5%, indicating modest spatial convergence. In particular, the fine-to-medium and medium-to-coarse mesh comparisons produce similar core-average differences in the range of 5.1%–5.55%. The spatial distribution of these errors is also non-random. The largest deviations are concentrated in localized regions of the inlet map, suggesting that mesh sensitivity is closely associated with the heterogeneous lower-plenum flow structures responsible for the inlet flow maldistribution.

For completeness, Appendix Appendix A provides complementary figures for the remaining axial layers, which exhibit more homogeneous flow distributions. One important observation can be drawn from the maximum-error plots in Figures A.11, Figures A.13, and A.15. Specifically, the fine-to-medium mesh comparison for Layers 2–9 in Figure A.11 produces significantly lower maximum-error averages than both the fine-to-coarse comparison in Figure A.13 and the medium-to-coarse comparison in Figure A.15.

Together, the axial flow maps and mesh-sensitivity results support the use of the high-fidelity mesh as the reference case for the machine-learning studies that follow. The high-fidelity simulation resolves the strongest inlet heterogeneity while providing a consistent basis for generating assembly-level training data. The mesh comparison also provides useful context for interpreting downstream data-driven results. Model errors near the base layer should be understood not only as a machine-learning challenge, but also as occurring in the same region where the CFD solution is most sensitive to spatial resolution.

These CFD results have several implications for reactor flow modeling and data-driven applications. First, they show that pump-induced swirl and lower-plenum transport can produce non-trivial assembly-level inlet conditions, supporting the need for high-fidelity CFD when generating reference data for reduced-order models. Second, the sharp axial reduction in flow heterogeneity indicates that the lower core region is likely the most challenging region for surrogate modeling, partial reconstruction, and transient prediction. This is consistent with the expectation that machine-learning models will have larger errors near the base plane, where the flow contains stronger gradients, more complex spatial structure, and greater mesh sensitivity. Third, the coherent spatial patterns in the assembly maps suggest that spatially aware learning methods, such as CNNs, ConvLSTMs, graph neural networks, and neural operators, are more appropriate than models that treat assembly responses as independent scalar time series.

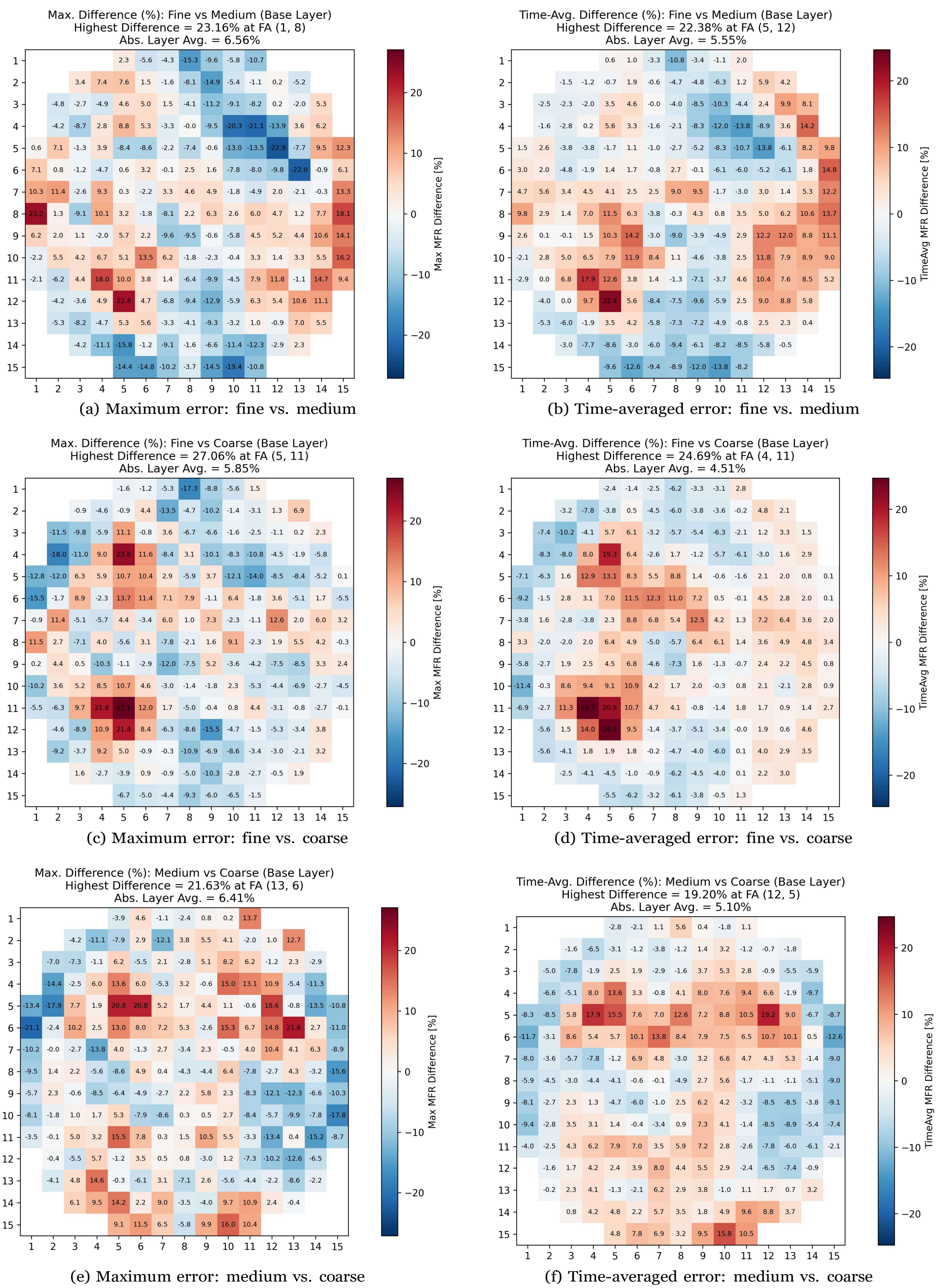

Figure 7: Mesh-sensitivity comparison of assembly-wise mass flow rate errors at the core inlet base layer. The left column shows the maximum percent error observed over the transient, while the right column shows the corresponding time-averaged percent error. Rows compare the fine and medium meshes, fine and coarse meshes, and medium and coarse meshes, respectively. “Abs. Layer Avg.” represents the absolute core-averaged error for the bottom layer, computed to avoid cancellation between positive and negative errors.

From a multiphysics perspective, the observed flow distribution motivates future coupling between the CFD-derived inlet fields and neutronics or subchannel calculations. Local variations in assembly coolant flow can influence moderator temperature, density feedback, and assembly-wise heat removal. While the present results do not by themselves quantify the neutronic impact of the flow asymmetry, they provide the spatially resolved inlet information needed to evaluate such effects in subsequent coupled analyses. In this sense, the CFD dataset serves not only as a stand-alone thermal-hydraulic result, but also as a foundation for multi-fidelity surrogate modeling, sparse sensing studies, and future reactor digital twin workflows.

## 5. Machine Learning Applications on the CFD Data

In this section, we present two distinct machine learning applications using the **high-fidelity simulation dataset** generated in this study. The first focuses on reconstructing the full-core flow field with a CNN, leveraging measurements from selected assemblies to infer conditions in unsensed regions. The second explores short-term autoregressive prediction of flow variables at the next time step using multiple approaches, including LSTM, ConvLSTM, and DeepONet. Finally, we discuss key limitations and research implications, highlighting opportunities for future development, building on the proposed models and dataset.

### *5.1. Flow Field Reconstruction with CNN*

We formulate flow-field reconstruction as a supervised inpainting problem on a multi-level reactor flow field. At each time step, the model receives partially observed assembly-level mass flow rates on a fixed $15 \times 15$ reactor-plane geometry and predicts the values at unobserved locations. In this work, four axial levels are reconstructed jointly so that the model can exploit both in-plane spatial structure and inter-level correlations. The learning objective is to recover the hidden values while exactly preserving the measurements that are already available.

Let $L = 4$ denote the number of reconstructed axial levels and $T$ the number of timesteps. For each level, the data are arranged as a time series of planar fields with shape $(T, H, W)$, where $(H, W) = (15, 15)$. A fixed geometry mask $\mathrm{M}_{\mathrm{geom}} \in \{0, 1\}^{H\times W}$ restricts the problem to physically meaningful assembly locations. To emulate partial instrumentation, we impose a synthetic missing-data pattern within the valid geometry using a checkerboard-style mask with 50% of valid cells hidden. The corresponding observed mask is defined as

$$\mathrm{M}_{\mathrm{obs}} = \mathrm{M}_{\mathrm{geom}} \wedge \neg \mathrm{M}_{\mathrm{miss}}.$$

Timesteps are then divided sequentially into training, validation, and test subsets using fractions of 0.45/0.10/0.45.

Normalization is performed independently for each axial level using only observed cells from the training split. If $x_\ell$ denotes the field at level $\ell$, the normalized quantity is

$$z_\ell = \frac{x_\ell - \mu_\ell}{\sigma_\ell}, \qquad \ell = 1, \ldots, L,$$

where $(\mu_\ell, \sigma_\ell)$ are computed from training observations only and then reused for validation and testing. This prevents information leakage and ensures that all reported errors can be mapped back to the physical scale after inference through de-normalization.

For each timestep, the 3D network input is constructed as

$$\mathbf{x} = [z \odot \mathrm{M}_{\mathrm{obs}}, \quad \mathrm{M}_{\mathrm{obs}}, \quad \mathrm{M}_{\mathrm{geom}}] \in \mathrm{R}^{3\times L\times H\times W},$$

where the three channels correspond to the observed values, the observed-mask indicator, and the geometry mask, respectively. The target tensor is

$$\mathbf{y} = z \odot \mathrm{M}_{\mathrm{geom}} \in \mathbb{R}^{1\times L\times H\times W}.$$

The missing mask is carried separately and is used to restrict both optimization and evaluation to intentionally hidden cells.

Reconstruction is performed using a fully convolutional 3D network consisting of an initial 3D convolutional stem, a stack of residual dilated 3D blocks, and a final $1 \times 1 \times 1$ convolutional head that produces a single reconstructed field. The dilation strategy is anisotropic. Dilation along the axial dimension is held fixed, while the in-plane dilations are varied to enlarge the lateral receptive field without distorting level-to-level alignment. Group normalization and Sigmoid Linear Unit (SiLU) activations are used throughout the network, with dropout applied in the residual blocks. Optional coordinate channels may also be concatenated to provide explicit positional information.

To guarantee that measured values are preserved, the final prediction is formed through an observed-value copy-through rule,

$$\hat{\mathbf{y}} = (z \odot \mathrm{M}_{\mathrm{obs}}) + \mathrm{M}_{\mathrm{miss}} \odot f_\theta(\mathbf{x}),$$

so that the network only learns to fill the missing region. Training minimizes a masked mean-squared error over hidden cells,

$$\mathcal{L} = \frac{\sum \left((\hat{\mathbf{y}} - \mathbf{y}) \odot \mathrm{M}_{\mathrm{miss}}\right)^2}{\sum \mathrm{M}_{\mathrm{miss}}},$$

using the AdamW optimizer with a scheduler to reduce the learning rate upon plateau. The checkpoint with minimum validation loss is retained for testing.

The complete reconstruction workflow is summarized in Figure 8. The figure connects the physical CFD domain, the axial sampling locations, the imposed partial-observation mask, and the resulting plane-wise reconstruction errors. This provides a direct visual link between the reactor geometry, the inpainting formulation, and the spatial structure of the model error.

Table 5: Plane-wise reconstruction performance on intentionally hidden assembly locations.

| Axial Plane | MAE (kg/s) | MAPE (%) | $R^2$ |
|---|---|---|---|
| Plane 1 | 2.6062 | 3.3473 | 0.7593 |
| Plane 2 | 0.0641 | 0.0825 | 0.9966 |
| Plane 3 | 0.0120 | 0.0154 | 0.9986 |
| Plane 4 | 0.0043 | 0.0055 | 0.9982 |

After training, predictions on the test set are de-normalized level by level and evaluated only over intentionally hidden cells. Reconstruction performance is reported for each axial level using mean absolute error (MAE), mean absolute percentage error (MAPE), and coefficient of determination ($R^2$), as summarized in Table 5. In addition to these scalar metrics, spatial MAE maps are generated to diagnose where reconstruction errors occur within the assembly grid.

As shown in Figure 8, the imposed checkerboard mask creates a deliberately partial reconstruction problem in which approximately half of the valid assembly locations are withheld. The quantitative results show a strong dependence on axial position. Plane 1 (base layer) has the largest reconstruction error, with an MAE of 2.6062 kg/s, MAPE of 3.3473%, and $R^2 = 0.7593$. In contrast, the downstream planes are reconstructed with substantially higher accuracy. Plane 2 achieves an MAE of 0.0641 kg/s and $R^2 = 0.9966$, while Planes 3 and 4 achieve MAE values of 0.0120 kg/s and 0.0043 kg/s, respectively, with $R^2$ values near 0.998.

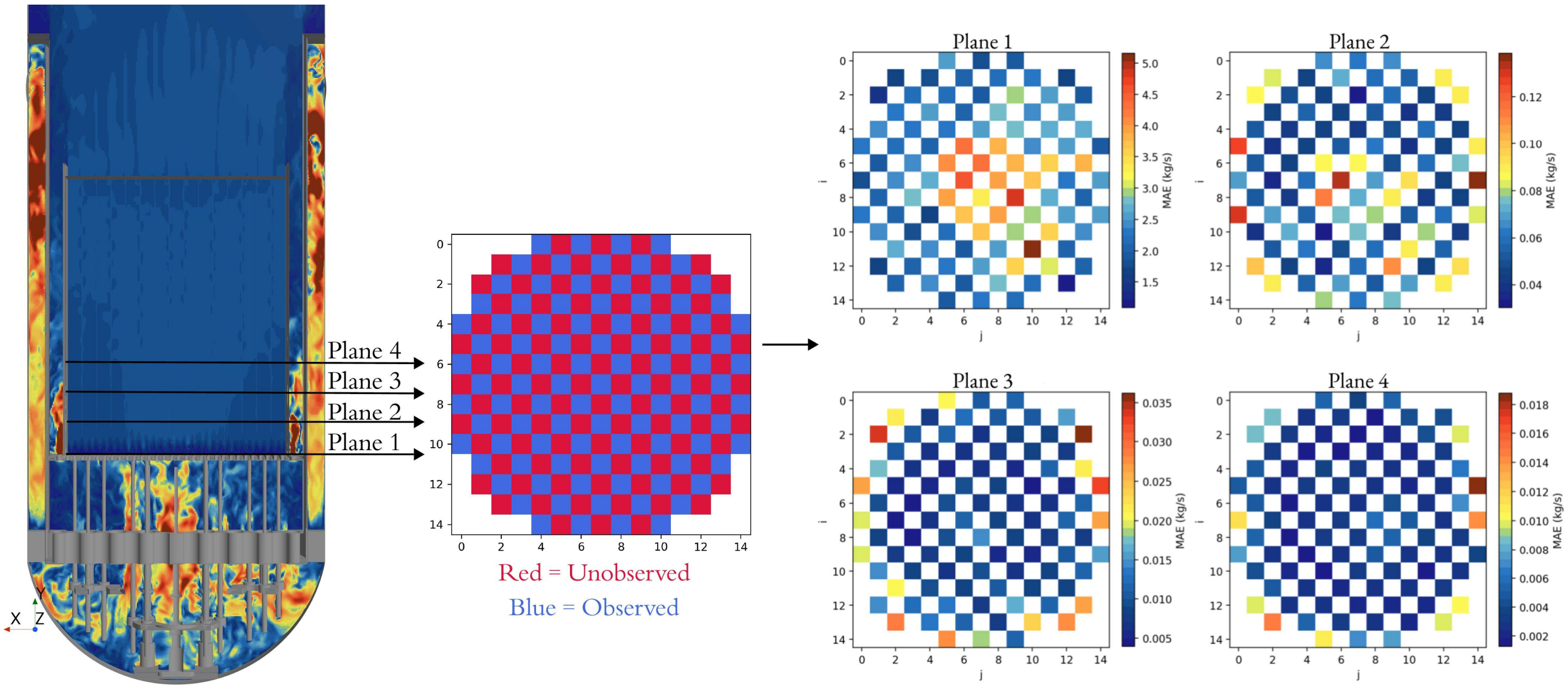

Figure 8: Partial CFD field reconstruction workflow and plane-wise error evaluation. A four-loop PWR CFD domain is sampled at four axial planes near the core inlet, where partial assembly-wise observations are imposed using a checkerboard mask. Blue cells denote observed assembly locations, while red cells denote unobserved locations to be reconstructed by the inpainting model. The right-hand panels show the plane-wise mean absolute error (MAE) distributions for the reconstructed mass flow rate field, indicating how reconstruction error varies spatially across the assembly grid and axially across the four sampled planes.

This trend is consistent with the spatial error maps in Figure 8. The reconstruction task is most difficult at Plane 1 (base layer), where the inlet flow field is more heterogeneous and contains stronger local assembly-to-assembly variation. Farther downstream, the flow becomes progressively more homogenized, reducing the difficulty of inferring missing assembly values from neighboring observations. The spatial MAE maps also show that the remaining errors are localized rather than uniformly distributed, with elevated errors occurring near regions where the partial observations provide weaker constraints on local flow structure. Therefore, the MAE maps complement the scalar metrics by identifying where reconstruction errors occur and by revealing how reconstruction difficulty changes across axial levels.

Hyperparameters were tuned in staged searches over optimization, architectural, and scheduler settings, and the final reported model corresponds to the configuration with the lowest validation loss. The training set comprised the first 4,500 transient snapshots of the flow corresponding to data taken from the [10, 12.25] second period of the transient. The validation set contains data from the [12.25, 12.75] second period (1,000 timesteps), and the test set contains the remaining 4,500 from the [12.75, 15] second period. Splitting in this fashion prevents the model from training on data that is more highly correlated in time with the test set and thus provides a more practical evaluation of the proposed field reconstruction framework.

### *5.2. One-Step Flow Field Prediction*

In recent years, scientific machine learning has increasingly incorporated neural operator frameworks, such as DeepONet [56], alongside traditional Computational Fluid Dynamics (CFD) approaches. In this work, we compare the performance of several machine learning models, including LSTM [57], ConvLSTM [58], and DeepONet. All models are trained using a single NVIDIA RTX PRO 6000 Blackwell GPU available on an internal server at the University of Michigan.

The dataset is structured for one-step-ahead prediction, where the model predicts the next state using the true immediately preceding state as input. This Markovian approach assumes that the immediate previous state contains sufficient information to evolve the system by one time step. To stabilize training, the data were normalized using a MinMaxScaler. The temporal sequence was partitioned into a 60/20/20 split for training, validation, and testing, respectively.

To capture the complex spatio-temporal dependencies, three architectures were implemented: LSTM, ConvLSTM, and DeepONet. The LSTM model consists of two stacked recurrent layers with a hidden dimension of 128, followed by a projection head with two hidden layers of 128 units each and GELU activation that maps the final hidden state to the target spatial grid. The ConvLSTM architecture preserves spatial topology by replacing standard LSTM inner products with 2D convolutional operations. It employs two ConvLSTM layers with 128 channels each and $3 \times 3$ kernels, followed by a convolutional output layer. DeepONet consists of branch and trunk networks, each with hidden layers of 128 and 256 units and GELU activation functions. The outputs are merged via an inner product across 128 units to reconstruct the field.

All models were trained using the AdamW optimizer to reduce overfitting. Training was conducted for 100 epochs using a Mean Squared Error (MSE) loss function. Hyperparameter tuning was performed, and the best configurations were selected for comparison.

From Table 6, the training results across the first three layers from the bottom of the fuel assembly (FA) highlight a clear progression in model performance as the flow develops. The "Base" layer remains the most challenging to predict, as observed in the previous application in Section 5.1, due to its high turbulent energy and strong spatio-temporal variability. In this region, both LSTM and DeepONet show noticeably lower accuracy (e.g., $R^2 = 0.4013$ and $0.5837$, respectively) and higher errors. As the flow becomes more structured in higher layers, all models improve, with LSTM and DeepONet achieving $R^2$ values near or above 0.9 in Layer 1, although some degradation is still observed in Layer 2 for DeepONet. In contrast, the ConvLSTM consistently maintains near-perfect accuracy across all layers ($R^2 \approx 1$) with orders-of-magnitude lower MAPE, demonstrating its strong capability to capture coupled spatial and temporal correlations. These results reinforce that purely temporal models, such as LSTM, and operator-learning approaches without explicit spatio-temporal coupling, such as DeepONet, struggle to resolve highly turbulent dynamics, whereas ConvLSTM effectively leverages spatial context to provide robust and generalizable predictions across varying flow regimes.

Table 6: Model performance comparison across the first three layers. The metrics are computed by comparing the predicted flow rates against the ground-truth high-fidelity simulation data over the final 2000 time steps of the test sequence. All metrics are evaluated on an unseen test set to assess the model's generalization performance.

| Layer | Metric | LSTM | ConvLSTM | DeepONet |
|---|---|---|---|---|
| Base | $R^2$ (-) | 0.4013 | **0.9999** | 0.5837 |
| | MAPE (%) | 4.803 | **0.07190** | 3.960 |
| Layer1 | $R^2$ (-) | 0.8998 | **1.000** | 0.9098 |
| | MAPE (%) | 0.4166 | **0.006000** | 0.3760 |
| Layer2 | $R^2$ (-) | 0.8771 | **0.9999** | 0.6224 |
| | MAPE (%) | 0.1591 | **0.004700** | 0.2511 |

In Figure 9, the time-averaged mass flow rate on the fuel assembly (FA) is compared across the first four layers using the ConvLSTM model. The results show excellent agreement between predictions and ground-truth distributions, with mean percentage differences that are negligibly small. The spatial error fields are generally low in magnitude, with slightly higher localized deviations observed in the Base layer, which is consistent with the higher turbulence intensity and complex inlet flow structures. As the flow develops downstream, the error becomes more uniformly distributed and reduced in magnitude, indicating improved predictability in more stable flow regions.

This behavior is further supported by the boxplot in Figure 10, which shows a wider spread and higher median error for the Base layer, with some outliers reaching higher absolute errors. In contrast, all subsequent layers exhibit tightly clustered distributions with significantly lower median and interquartile ranges, confirming that the model achieves consistently low errors once the flow stabilizes. The decreasing variance across layers demonstrates that the ConvLSTM architecture effectively captures the dominant spatio-temporal correlations while maintaining robustness against temporal fluctuations. Overall, these results highlight that the model not only preserves the global flow structure but also improves reliability in regions with more developed and less chaotic flow dynamics.

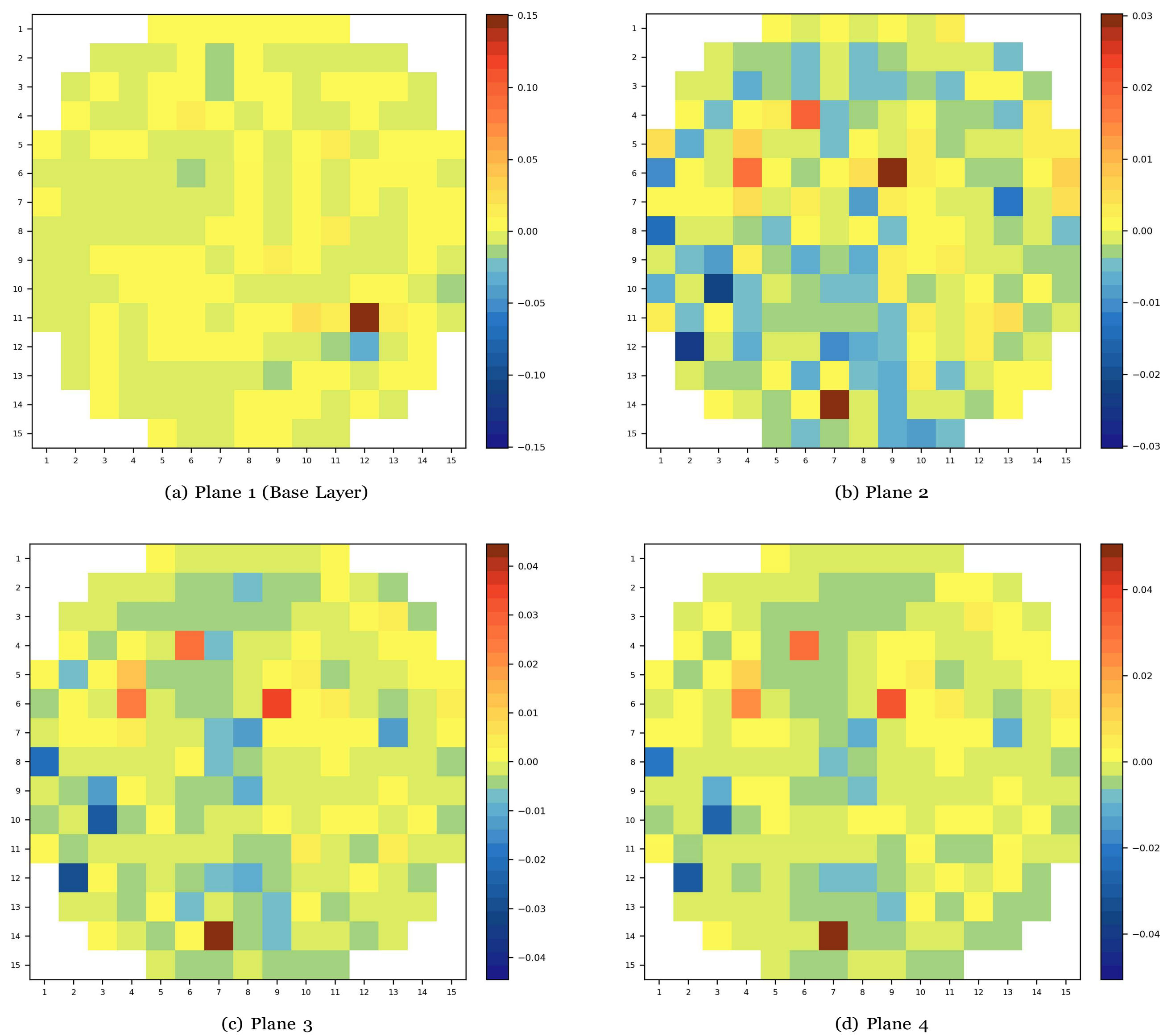

(a) Plane 1 (Base Layer)

(b) Plane 2

(c) Plane 3

(d) Plane 4

Figure 9: Assembly-wise distributions of the mean absolute percentage error (MAPE, %) across the first four axial layers are shown for the best-performing ConvLSTM autoregressive model. The error is computed by comparing the predicted flow rates against the ground-truth high-fidelity simulation data over the final 2000 time steps of the test sequence. All metrics are evaluated on an unseen test set to assess the model's generalization performance.

### *5.3. Research Implications*

The results of this study highlight several key implications for advancing data-driven modeling of reactor thermal-hydraulics, offering clear directions for future work and potential extensions by the broader scientific

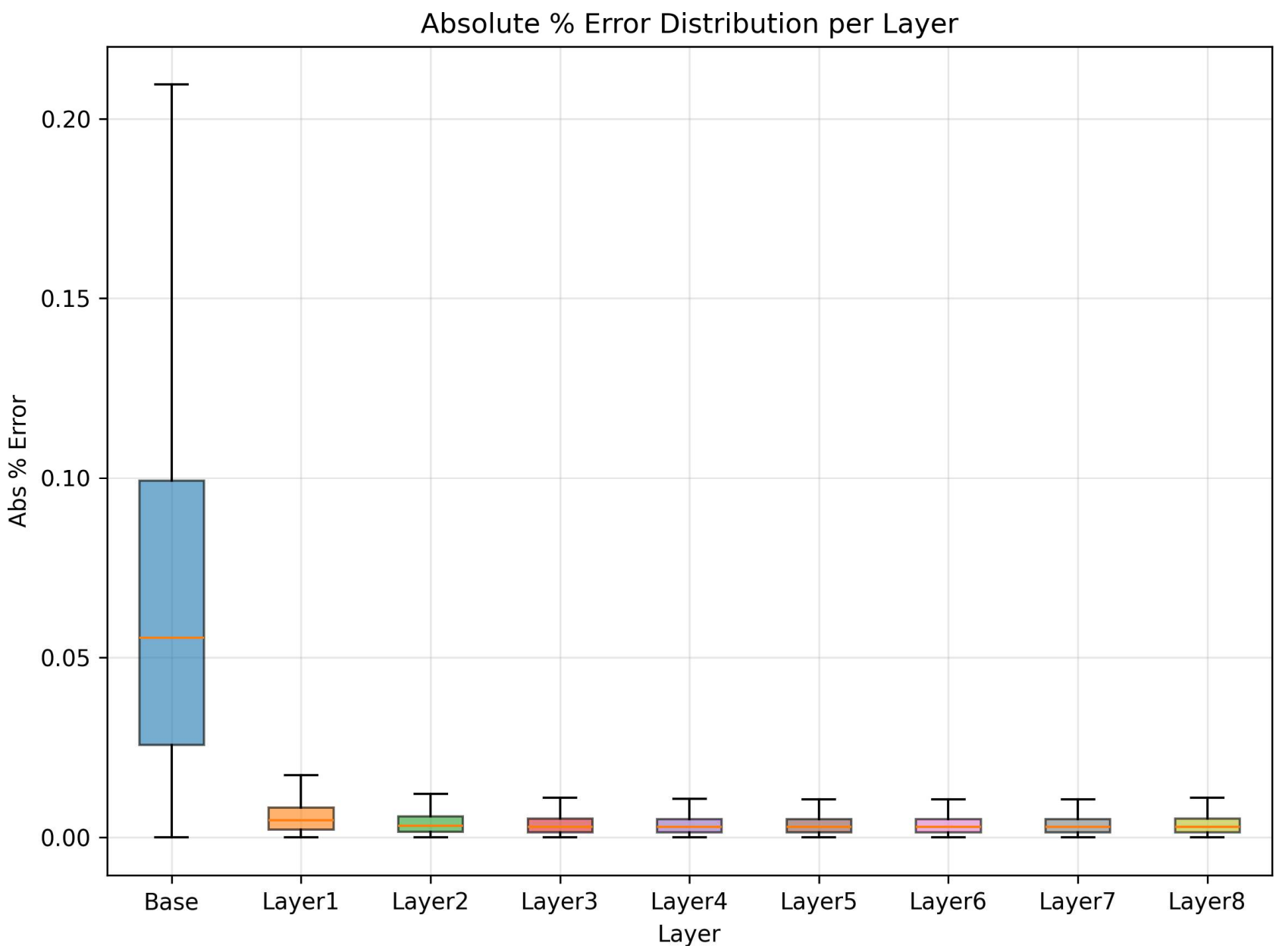

Figure 10: Box plots of the absolute percentage error distribution (e.g., MAPE) across the nine axial layers, based on the best-performing ConvLSTM autoregressive model. The metrics are evaluated on an unseen test set derived from the high-fidelity simulation model.

community.

First, the strong performance of ConvLSTM highlights the growing importance of GPU-accelerated computing for training and deploying spatio-temporal machine learning models. As datasets from high-fidelity CFD simulations continue to grow in size and complexity, efficient GPU utilization becomes essential to enable scalable training and potentially real-time inference in applications such as digital twins and online monitoring systems. This requirement extends beyond ConvLSTM to more advanced architectures, including neural operators (e.g., Transolvers) and diffusion-based models, which often demand significantly higher computational cost and training time. As such, future developments in this space will increasingly rely on access to high-performance computing infrastructure to support both model training and deployment.

Second, the spatial structure observed in the assembly-level flow fields suggests that graph-based learning approaches could be highly effective. Graph Neural Networks (GNNs) offer a natural representation of reactor cores as interconnected assemblies, where physical adjacency and coupling can be explicitly encoded. This may provide improved generalization compared to grid-based convolutional models, especially for irregular geometries or partially observed systems.

Third, the strong dependence of prediction accuracy on axial location—especially the difficulty associated with the highly turbulent base layer—combined with the substantial computational cost of transient simulations (on the order of 6–20 days), underscores the need for multi-fidelity modeling strategies. Integrating low-, medium-, and high-fidelity simulations within a unified framework can enhance robustness while significantly reducing computational expense. Such approaches enable the use of inexpensive coarse simulations that are systematically corrected using high-fidelity data, ultimately leading to more efficient and scalable surrogate models.

Fourth, the observed spatial heterogeneity of the inlet flow suggests that clustering-based approaches could improve learning performance. By grouping assemblies with similar flow characteristics prior to training, models can focus on learning localized flow behaviors rather than attempting to capture all regimes simultaneously. Such clustering-then-learning strategies may reduce model complexity and improve accuracy, particularly in re-gions with strong gradients and turbulence. Fourth, the observed spatial heterogeneity of the inlet flow suggests that clustering-based approaches could improve learning performance. By grouping assemblies with similar flow characteristics prior to training, models can focus on learning localized flow behaviors rather than attempting to capture all regimes simultaneously. Such clustering-then-learning strategies may reduce model complexity and improve accuracy, particularly in regions with strong gradients and turbulence. Methods such as principal component analysis (PCA), proper orthogonal decomposition (POD), or more advanced techniques rooted in graph theory may be leveraged for this purpose to identify dominant flow structures and guide the clustering process.

Fifth, the use of machine learning in safety-critical nuclear applications necessitates the integration of explainable AI (XAI) techniques. While the models presented in this work achieve high predictive accuracy, their interpretability remains limited. Incorporating methods such as saliency analysis, feature attribution, or physics-informed interpretability could provide insight into how models make predictions, thereby increasing trust and supporting regulatory acceptance. Examples of these methods will include interpretable machine learning models and post-hoc explainability techniques such as DeepLIFT, both of which have been previously evaluated in nuclear engineering applications [59, 60].

Finally, the results highlight a strong opportunity for coupling with neutronics and other multiphysics models. Assembly-level variations in mass flow rate can directly impact moderator temperature, density feedback, and reactivity distributions. Integrating the present thermal-hydraulic predictions with neutronic solvers would enable more comprehensive analyses of the reactor power behavior in the BEAVRS core and could support advanced multiphysics digital twin frameworks.

Despite these promising implications, several limitations must be acknowledged. Most notably, the CFD simulations used in this study have not been directly validated against full-scale experimental data. Obtaining detailed, assembly-level flow measurements in operating reactor systems remains extremely challenging due to instrumentation limitations and restricted industry access. As a result, the present work relies on high-fidelity simulations as a surrogate for ground truth.

A second limitation is associated with the mesh sensitivity results shown in Figure 7, where full spatial convergence was not achieved in the bottom layer. Both the time-averaged and maximum error metrics indicate elevated discrepancies in this region, which is consistent with the highly turbulent and heterogeneous inlet conditions. In this study, computational resource constraints limited the ability to run longer transient simulations that would allow for additional flow development, mixing, and stabilization. This highlights an important direction for future work. Rather than initiating data collection after approximately 10 seconds of transient evolution, future studies could extend simulations to longer durations (e.g., 60 seconds) before sampling, which is expected to improve both spatial and temporal convergence of the solution.

Nevertheless, the results remain consistent with expected physical behavior, including strong inlet heterogeneity, lower-plenum mixing effects, and axial flow homogenization. This agreement with established thermal-hydraulic principles provides confidence in the qualitative validity of the findings.

Future work in CFD modeling should focus on incorporating experimental benchmarks, reduced-scale test data, or plant measurements to further validate the simulations. In addition, leveraging GPU-accelerated CFD capabilities, such as those available in STAR-CCM+, could enable significantly longer and more detailed simulations. Recent developments suggest that GPU-based CFD can achieve speedups on the order of 3×–4× compared to traditional CPU-based approaches, which would directly support improved convergence studies and higher-fidelity data generation for both CFD and machine learning applications.

## 6. Conclusions

This work develops a high-fidelity CFD modeling and data-driven analysis pipeline for assembly-level flow characterization in a four-loop pressurized water reactor. Using publicly available geometric and operating data, a full lower-plenum and core-inlet flow domain was constructed and used to generate transient CFD datasets with pump-induced swirl boundary conditions. The resulting simulations show that cold-leg swirl and lower-plenum transport produce strongly heterogeneous assembly-wise inlet flow distributions, especially near the lower core region, while axial resistance and mixing through the fuel assembly region progressively homogenize the flow at higher elevations. These spatially structured CFD fields were then used to demonstrate machine learning applications for partial field reconstruction and transient prediction. The 3D CNN inpainting model accurately reconstructed missing assembly-level mass flow values from partial checkerboard observations, with the largest errors occurring near the most heterogeneous inlet plane and substantially smaller errors in downstream planes. The broader ML comparison further indicates that models preserving spatial structure, particularly ConvLSTM-based approaches, are better suited for predicting these reactor flow fields than models that treat the response as weakly coupled scalar time series. Overall, the results support the use of high-fidelity CFD datasets as reference data for reduced-order modeling, sparse sensing, surrogate modeling, and future multiphysics coupling workflows. This research also opens a variety of potential future work extensions, which were outlined in detail in Section 5.3.

### Data Availability

The authors have all the datasets and codes to reproduce all the results in this work currently in a private GitHub repository. To ensure the confidentiality of this research, the authors will make this repository public during an advanced stage of the review process, which will be listed under our research group's public Github page: https://github.com/aims-umich. Given the significant storage demands, the raw STAR-CCM+ models will be provided by the authors upon request. Nevertheless, the geometry CAD files will be made available on that GitHub repository.

### Acknowledgment

The authors gratefully acknowledge Alessandro Persico for his valuable insights and contributions to this work through regular monthly meetings. This research is sponsored by the Department of Energy Office of Nuclear Energy through the Nuclear Energy University Programs (Award number: DE-NE0009506). The first author (L. A. Burnett) received sponsorship through the National Science Foundation's Graduate Research Fellowship Program (Grant Number: DGE 2241144).

### CRediT Author Statement

**Logan A. Burnett**: Conceptualization, Methodology, Software, Validation, Formal Analysis, Visualization, Investigation, Data Curation, Writing - Original Draft; **Hyungjun Kim**: Methodology, Software, Validation, Formal Analysis, Visualization, Writing - Original Draft; **Hsien-Cheng Chou**: Methodology, Software, Validation, Data Curation, Visualization, Writing - Review and Edit; **Arsha Witoelar**: Methodology, Software, Validation, Data Curation, Visualization, Writing - Review and Edit; **Robert A. Brewster**: Conceptualization, Methodology, Software, Validation, Writing - Review and Edit; **Benoit Forget**: Conceptualization, Funding Acquisition, Project Administration, Writing - Review and Edit; **Emilio Baglietto**: Conceptualization, Methodology, Funding Acquisition, Supervision, Project Administration, Writing - Review and Edit; **Majdi I. Radaideh**: Conceptualization, Methodology, Resources, Funding Acquisition, Supervision, Project Administration, Writing - Original Draft.

## References

[1] N. E. Todreas, M. S. Kazimi, Nuclear Systems Volume I: Thermal Hydraulic Fundamentals, CRC Press, 2021.

[2] M. Wang, Y. Wang, W. Tian, S. Qiu, G. H. Su, Recent progress of CFD applications in PWR thermal hydraulics study and future directions, Annals of Nuclear Energy 150 (2021) 107836. doi:10.1016/j.anucene.2020.107836.

[3] J. Fang, M. Tano, N. Saini, A. Tomboulides, V. Coppo-Leite, E. Merzari, B. Feng, D. Shaver, CFD simulations of molten salt fast reactor core cavity flows, Nuclear Engineering and Design 424 (2024) 113294. doi:10.1016/j.nucengdes.2024.113294.

[4] R. Freile, M. Tano, P. Balestra, S. Schunert, M. Kimber, Improved natural convection heat transfer correlations for reactor cavity cooling systems of high-temperature gas-cooled reactors: From computational fluid dynamics to Pronghorn, Annals of Nuclear Energy 163 (2021) 108547. doi:10.1016/j.anucene.2021.108547.

[5] C. F. Boyd, D. M. Helton, K. Hardesty, CFD analysis of full-scale steam generator inlet plenum mixing during a PWR severe accident, Division of Systems Analysis and Regulatory Effectiveness, Office of Nuclear Regulatory Research, U.S. Nuclear Regulatory Commission, 2004.

[6] M. Pham, G. Bois, F. Francois, E. Baglietto, Assessment of state-of-the-art multiphase CFD modeling for subcooled flow boiling in reactor applications, Nuclear Engineering and Design 411 (2023) 112379. doi:10.1016/j.nucengdes.2023.112379.

[7] T. Wei, B. Zhang, S. Wang, S. Tan, D. Li, S. Qiao, Numerical analysis of passive safety injection driven by natural circulation in floating nuclear power plant, Energy 263 (2023) 126077. doi:10.1016/j.energy.2022.126077.

[8] R. Hu, L. Zou, D. O'Grady, T. Mui, Z. J. Ooi, G. Hu, E. Cervi, G. Yang, D. Andrs, A. Lindsay, et al., SAM: A modern system code for advanced non-LWR safety analysis, Nuclear Technology 211 (9) (2025) 1883–1902.

[9] R. Salko, A. Wysocki, T. Blyth, A. Toptan, J. Hu, V. Kumar, C. Dances, W. Dawn, Y. Sung, V. Kucukboyaci, et al., CTF: A modernized, production-level, thermal hydraulic solver for the solution of industry-relevant challenge problems in pressurized water reactors, Nuclear Engineering and Design 397 (2022) 111927. doi:10.1016/j.nucengdes.2022.111927.

[10] C. Boyd, Perspectives on CFD analysis in nuclear reactor regulation, Nuclear Engineering and Design 299 (2016) 12–17. doi:10.1016/j.nucengdes.2015.11.033.

[11] Y. Liu, D. Wang, X. Sun, N. Dinh, R. Hu, Uncertainty quantification for multiphase-CFD simulations of bubbly flows: A machine learning-based Bayesian approach supported by high-resolution experiments, Reliability Engineering & System Safety 212 (2021) 107636. doi:10.1016/j.ress.2021.107636.

[12] Y. Liao, CFD modelling of flashing flows for nuclear safety analysis: Possibilities and challenges, Kerntechnik 89 (2) (2024) 169–184.

[13] T. Höhne, E. Krepper, U. Rohde, Application of CFD codes in nuclear reactor safety analysis, Science and Technology of Nuclear Installations 2010 (1) (2010) 198758. doi:10.1155/2010/198758.

[14] V. Petrov, A. Manera, Effect of pump-induced cold-leg swirls on the flow field in the RPV of the EPR™: CFD investigations and comparison with experimental results, Nuclear Engineering and Design 241 (5) (2011) 1478–1485. doi:10.1016/j.nucengdes.2011.01.040.

[15] R. W. Johnson, R. R. Schultz, P. J. Roache, I. B. Celik, W. D. Pointer, Y. A. Hassan, Processes and procedures for application of CFD to nuclear reactor safety analysis, Tech. rep., Idaho National Laboratory, Idaho Falls, ID, United States (2006).

[16] D. Bertolotto, A. Manera, S. Frey, H.-M. Prasser, R. Chawla, Single-phase mixing studies by means of a

directly coupled CFD/system-code tool, Annals of Nuclear Energy 36 (3) (2009) 310–316. doi:10.1016/j.anucene.2008.12.017.

[17] M. E. Conner, E. Baglietto, A. M. Elmahdi, CFD methodology and validation for single-phase flow in PWR fuel assemblies, Nuclear Engineering and Design 240 (9) (2010) 2088–2095. doi:10.1016/j.nucengdes.2009.12.031.

[18] B. Han, X. Zhu, B.-W. Yang, S. Liu, A. Liu, Verification and validation of CFD and its application in PWR fuel assembly, Progress in Nuclear Energy 154 (2022) 104485. doi:10.1016/j.pnucene.2022.104485.

[19] M. Wang, L. Wang, X. Wang, J. Ge, W. Tian, S. Qiu, G. H. Su, CFD simulation on the flow characteristics in the PWR lower plenum with different internal structures, Nuclear Engineering and Design 364 (2020) 110705. doi:10.1016/j.nucengdes.2020.110705.

[20] Siemens Digital Industries Software, Simcenter STAR-CCM+, version 2502.0001 (2025).
URL https://www.plm.automation.siemens.com/global/en/products/simcenter/STAR-CCM.html

[21] M. Wang, H. Ju, J. Wu, H. Qiu, K. Liu, W. Tian, G. H. Su, A review of CFD studies on thermal hydraulic analysis of coolant flow through fuel rod bundles in nuclear reactor, Progress in Nuclear Energy 171 (2024) 105175. doi:10.1016/j.pnucene.2024.105175.

[22] J. Feng, L. Xu, E. Baglietto, Assessing the applicability of the Structure-Based Turbulence Resolution Approach to nuclear safety-related issues, Journal of Nuclear Engineering 2 (1) (2021) 61–85. doi:10.3390/jne2010005.

[23] M. Pham, V. Petrov, A. Manera, E. Baglietto, Assessing the Structure-Based Turbulence Model performance for thermal striping applications using symmetric jet experiments, Nuclear Technology 210 (7) (2024) 1212–1222. doi:10.1080/00295450.2023.2204989.
URL https://doi.org/10.1080/00295450.2023.2204989

[24] L. Harbour, G. Giudicelli, A. D. Lindsay, P. German, J. Hansel, C. Icenhour, M. Li, J. M. Miller, R. H. Stogner, P. Behne, D. Yankura, Z. M. Prince, C. DeChant, D. Schwen, B. W. Spencer, M. Tano, N. Choi, Y. Wang, M. Nezdyur, Y. Miao, T. Hu, S. Kumar, C. Matthews, B. Langley, N. Nobre, A. Blair, C. MacMackin, H. B. Rocha, E. Palmer, J. Carter, J. Meier, A. E. Slaughter, D. Andrˇs, R. W. Carlsen, F. Kong, D. R. Gaston, C. J. Permann, 4.0 MOOSE: Enabling massively parallel Multiphysics simulation, SoftwareX 31 (2025) 102264. doi:10.1016/j.softx.2025.102264.
URL https://www.sciencedirect.com/science/article/pii/S2352711025002316

[25] L. Zou, D. Nunez, R. Hu, Development and validation of SAM multi-dimensional flow model for thermal mixing and stratification modeling, Tech. rep., Argonne National Laboratory, Argonne, IL, United States (Jun. 2020). doi:10.2172/1671335.
URL https://www.osti.gov/biblio/1671335

[26] D. Price, N. Roskoff, M. I. Radaideh, B. Kochunas, Thermal modeling of an eVinci™-like heat pipe microreactor using OpenFOAM, Nuclear Engineering and Design 415 (2023) 112709. doi:10.1016/j.nucengdes.2023.112709.

[27] D. Price, N. Roskoff, M. I. Radaideh, B. Kochunas, Multiphysics modeling of heat pipe microreactor with critical control drum position search, Nuclear Science and Engineering (2024) 1–20.

[28] P. Wang, W. Liang, H. Gong, J. Chen, Decoupling control of core power and axial power distribution for large pressurized water reactors based on reinforcement learning, Energy 313 (2024) 133967.

[29] P. Wang, J. Zhang, J. Wan, S. Wu, A fault diagnosis method for small pressurized water reactors based on long short-term memory networks, Energy 239 (2022) 122298.

[30] R. Luo, Y. Li, H. Guo, Q. Wang, X. Wang, Cross-operating-condition fault diagnosis of a small module reactor based on cnn-lstm transfer learning with limited data, Energy 313 (2024) 133901.

[31] X. Yuan, T. Bai, C. Peng, Hybrid modeling method for reactor coolant loop combining data-driven and physics-based constraints, Energy (2026) 140177.

[32] X. Guo, W. Li, F. Iorio, Convolutional neural networks for steady flow approximation, in: Proceedings of the 22nd ACM SIGKDD International Conference on Knowledge Discovery and Data Mining, KDD '16, Association for Computing Machinery, New York, NY, USA, 2016, pp. 481–490. doi:10.1145/2939672.2939738.
URL https://doi.org/10.1145/2939672.2939738

[33] Y. Liu, R. Hu, A. Kraus, P. Balaprakash, A. Obabko, Data-driven modeling of coarse mesh turbulence for reactor transient analysis using convolutional recurrent neural networks, Nuclear Engineering and Design 390 (2022) 111716. doi:10.1016/j.nucengdes.2022.111716.
URL https://www.sciencedirect.com/science/article/pii/S002954932200070X

[34] L. Guastoni, A. Güemes, A. Ianiro, S. Discetti, P. Schlatter, H. Azizpour, R. Vinuesa, Convolutional-network models to predict wall-bounded turbulence from wall quantities, Journal of Fluid Mechanics 928 (2021) A27. doi:10.1017/jfm.2021.812.

[35] K. Stylianos, L. He, L. Jiao, O. Mehdizadeh, A. Garner, H. Dou, R. P. Taleyarkhan, J. Tan, Physics-informed neural network with transfer learning (TL-PINN) based on domain similarity measure for prediction of nuclear reactor transients, Scientific Reports 13 (1) (2023) 16840. doi:10.1038/s41598-023-43325-1.

[36] M. I. Radaideh, C. Pigg, T. Kozlowski, Y. Deng, A. Qu, Neural-based time series forecasting of loss of coolant accidents in nuclear power plants, Expert Systems with Applications 160 (2020) 113699. doi:10.1016/j.eswa.2020.113699.

[37] M. I. Radaideh, T. Kozlowski, Combining simulations and data with deep learning and uncertainty quantification for advanced energy modeling, International Journal of Energy Research 43 (14) (2019) 7866–7890.

[38] M. I. Radaideh, T. Kozlowski, Analyzing nuclear reactor simulation data and uncertainty with the group method of data handling, Nuclear Engineering and Technology 52 (2) (2020) 287–295.

[39] M. I. Radaideh, T. Kozlowski, Surrogate modeling of advanced computer simulations using deep Gaussian processes, Reliability Engineering & System Safety 195 (2020) 106731. doi:10.1016/j.ress.2019.106731.

[40] R. A. Saleem, M. I. Radaideh, T. Kozlowski, Application of deep neural networks for high-dimensional large bwr core neutronics, Nuclear Engineering and Technology 52 (12) (2020) 2709–2716.

[41] M. Eaheart, J. Cooper, M. Ross, N. See, M. I. Radaideh, Sensitivity analysis and uncertainty propagation of the time to onset of natural circulation in air ingress accidents, Nuclear Engineering and Design 445 (2025) 114510.

[42] D. Price, M. I. Radaideh, B. Kochunas, Multiobjective optimization of nuclear microreactor reactivity control system operation with swarm and evolutionary algorithms, Nuclear Engineering and Design 393 (2022) 111776.

[43] M. I. Radaideh, L. Tunkle, D. Price, K. Abdulraheem, L. Lin, M. Elias, Multistep criticality search and power shaping in nuclear microreactors with deep reinforcement learning, Nuclear Science and Engineering 200 (sup1) (2026) S309–S321.

[44] M. Eaheart, M. I. Radaideh, Multifidelity surrogate modeling of depressurized loss of forced cooling in high-temperature gas reactors, arXiv preprint arXiv:2603.14143.

[45] U. M. Nabila, L. Lin, X. Zhao, W. L. Gurecky, P. Ramuhalli, M. I. Radaideh, Data efficiency assessment of generative adversarial networks in energy applications, Energy and AI 20 (2025) 100501.

[46] L. A. Burnett, U. M. Nabila, M. I. Radaideh, Variational digital twins, Energy and AI 24 (2026) 100756.

[47] U. M. Nabila, P. Seurin, L. Lin, M. I. Radaideh, Physics-informed digital twin development of thermal energy distribution systems with active learning, Applied Energy 416 (2026) 127903.

[48] N. Horelik, B. Herman, B. Forget, K. Smith, Benchmark for evaluation and validation of reactor simulations (BEAVRS), v1.0.1, in: Proceedings of the International Conference on Mathematics and Computational Methods Applied to Nuclear Science and Engineering, Vol. 7, 2013, pp. 63–68.

[49] R. Ellis, System definition document: Reactor data necessary for modeling plutonium disposition in Catawba nuclear station units 1 and 2, Tech. rep., Oak Ridge National Laboratory (Nov. 2000). doi:10.2172/769301.

[50] United States Nuclear Regulatory Commission, Westinghouse technology systems manual (ML20057E160) (Jul. 2020).
URL https://www.nrc.gov/docs/ML2005/ML20057E160.pdf

[51] R. Wiser, E. Baglietto, J. Schneider, M. H. Anderson, Validation of URANS and STRUCT-$\varepsilon$ turbulence models for stratified sodium flow, Nuclear Engineering and Design 399 (2022) 112009. doi:10.1016/j.nucengdes.2022.112009.
URL https://www.sciencedirect.com/science/article/pii/S0029549322003600

[52] E. Baglietto, H. Ninokata, Anisotropic eddy viscosity modeling for application to industrial engineering internal flows, International Journal of Transport Phenomena 8 (2) (2006) 109–122.

[53] G. Lenci, E. Baglietto, A structure-based approach for topological resolution of coherent turbulence: Overview and demonstration, 16th International Topical Meeting on Nuclear Reactor Thermal Hydraulics (2015) 1–14.

[54] E. Baglietto, G. Lenci, D. Concu, STRUCT: A second-generation URANS approach for effective design of advanced systems, in: Proceedings of the Fluids Engineering Division Summer Meeting, Vol. Volume 1B of Fluids Engineering Division Summer Meeting, 2017, p. V01BT12A004. doi:10.1115/FEDSM2017-69241.
URL https://doi.org/10.1115/FEDSM2017-69241

[55] J. Feng, E. Baglietto, K. Tanimoto, Y. Kondo, Demonstration of the STRUCT turbulence model for mesh consistent resolution of unsteady thermal mixing in a T-junction, Nuclear Engineering and Design 361 (2020) 110572. doi:10.1016/j.nucengdes.2020.110572.
URL https://www.sciencedirect.com/science/article/pii/S0029549320300674

[56] L. Lu, P. Jin, G. Pang, Z. Zhang, G. E. Karniadakis, Learning nonlinear operators via DeepONet based on the universal approximation theorem of operators, Nature Machine Intelligence 3 (3) (2021) 218–229. doi:10.1038/s42256-021-00302-5.
URL https://doi.org/10.1038/s42256-021-00302-5

[57] S. Hochreiter, J. Schmidhuber, Long short-term memory, Neural Computation 9 (8) (1997) 1735–1780. doi:10.1162/neco.1997.9.8.1735.

[58] X. Shi, Z. Chen, H. Wang, D.-Y. Yeung, W.-K. Wong, W.-c. Woo, Convolutional LSTM network: A machine learning approach for precipitation nowcasting, in: Advances in Neural Information Processing Systems, 2015, pp. 802–810.

[59] P. A. Myers, N. Panczyk, S. Chidige, C. Craig, J. Cooper, V. Joynt, M. I. Radaideh, pymaise: A python platform for automatic machine learning and accelerated development for nuclear power applications, Progress in Nuclear Energy 180 (2025) 105568.

[60] N. R. Panczyk, O. F. Erdem, M. I. Radaideh, Opening the ai black-box: Symbolic regression with kolmogorov-arnold networks for advanced energy applications, Energy and AI (2025) 100595.

## Appendix A. Flow Error Results for all axial layers

The results presented in this appendix (Figures A.11–A.16) complement those in Figure 7, illustrating the maximum time-average flow errors across all axial layers—including the base layer at the core inlet—when comparing the three different models: fine, medium, and coarse.

Max. Difference (%): Fine vs Medium (9 Layers)
Global Highest Difference = 23.16% at Base Layer, FA (1, 8)

Base Layer
Highest Difference = 23.16% at FA (1, 8)
Abs. Layer Avg. = 6.56%

Layer 1
Highest Difference = 4.35% at FA (15, 11)
Abs. Layer Avg. = 1.50%

Layer 2
Highest Difference = -1.70% at FA (8, 15)
Abs. Layer Avg. = 0.55%

Layer 3
Highest Difference = 0.89% at FA (14, 8)
Abs. Layer Avg. = 0.27%

Layer 4
Highest Difference = 0.65% at FA (14, 11)
Abs. Layer Avg. = 0.19%

Layer 5
Highest Difference = 0.59% at FA (14, 5)
Abs. Layer Avg. = 0.18%

Layer 6
Highest Difference = -0.77% at FA (4, 9)
Abs. Layer Avg. = 0.21%

Layer 7
Highest Difference = -1.32% at FA (6, 12)
Abs. Layer Avg. = 0.26%

Layer 8
Highest Difference = -2.95% at FA (7, 7)
Abs. Layer Avg. = 0.51%

Figure A.11: Maximum error: Fine vs Medium mesh comparison for all axial layers.

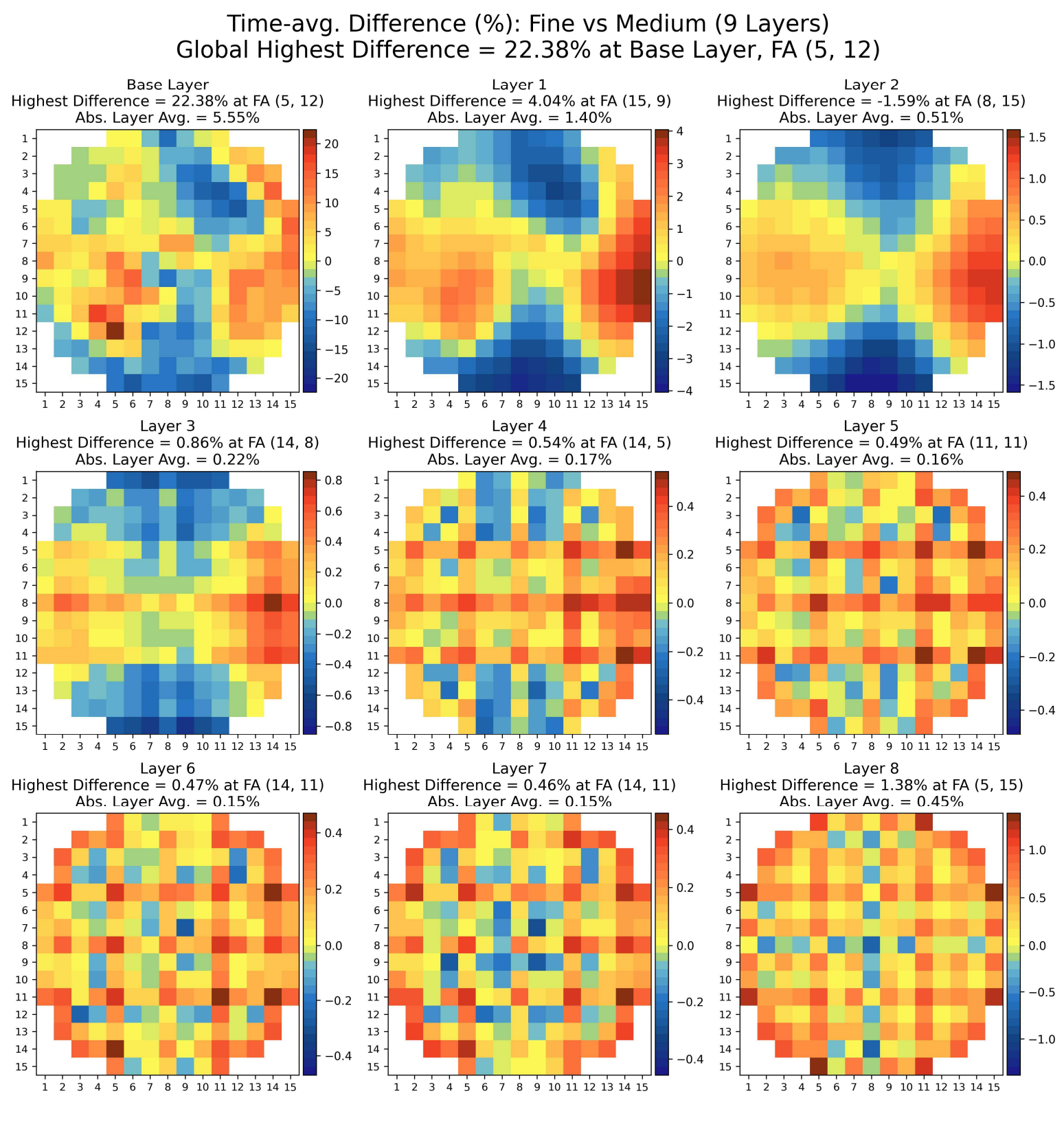

Figure A.12: Time-averaged error: Fine vs Medium mesh comparison for all axial layers.

Max. Difference (%): Fine vs Coarse (9 Layers)
Global Highest Difference = 27.06% at Base Layer, FA (5, 11)

Base Layer
Highest Difference = 27.06% at FA (5, 11)
Abs. Layer Avg. = 5.85%

Layer 1
Highest Difference = -3.21% at FA (8, 1)
Abs. Layer Avg. = 0.88%

Layer 2
Highest Difference = -14.91% at FA (5, 14)
Abs. Layer Avg. = 1.80%

Layer 3
Highest Difference = -20.24% at FA (4, 8)
Abs. Layer Avg. = 2.21%

Layer 4
Highest Difference = -24.31% at FA (3, 9)
Abs. Layer Avg. = 1.61%

Layer 5
Highest Difference = -19.62% at FA (10, 15)
Abs. Layer Avg. = 1.97%

Layer 6
Highest Difference = -18.10% at FA (11, 14)
Abs. Layer Avg. = 1.65%

Layer 7
Highest Difference = -22.13% at FA (2, 11)
Abs. Layer Avg. = 1.66%

Layer 8
Highest Difference = -15.24% at FA (13, 11)
Abs. Layer Avg. = 1.39%

Figure A.13: Maximum error: Fine vs Coarse mesh comparison for all axial layers.

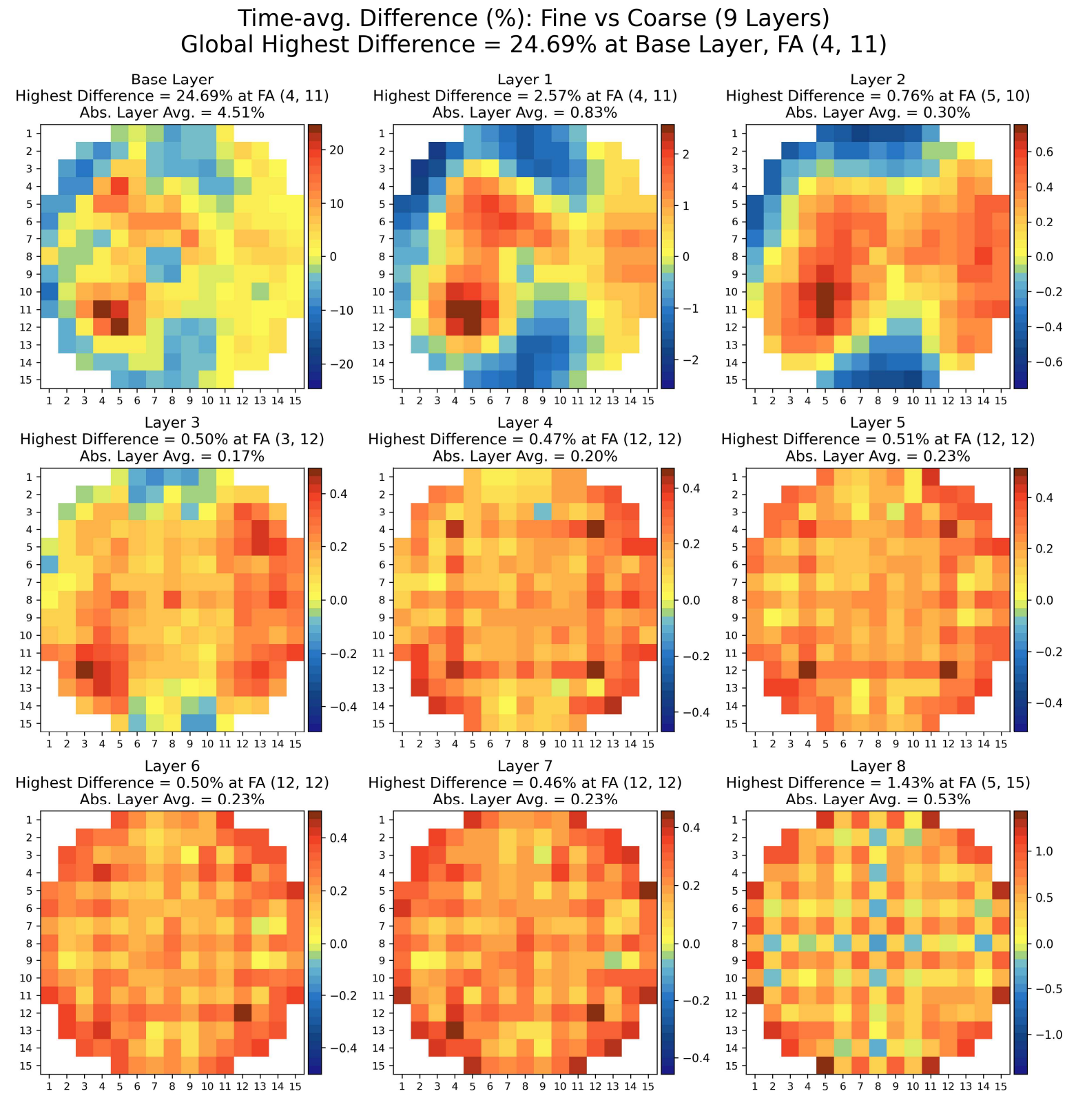

Figure A.14: Time-averaged error: Fine vs Coarse mesh comparison for all axial layers.

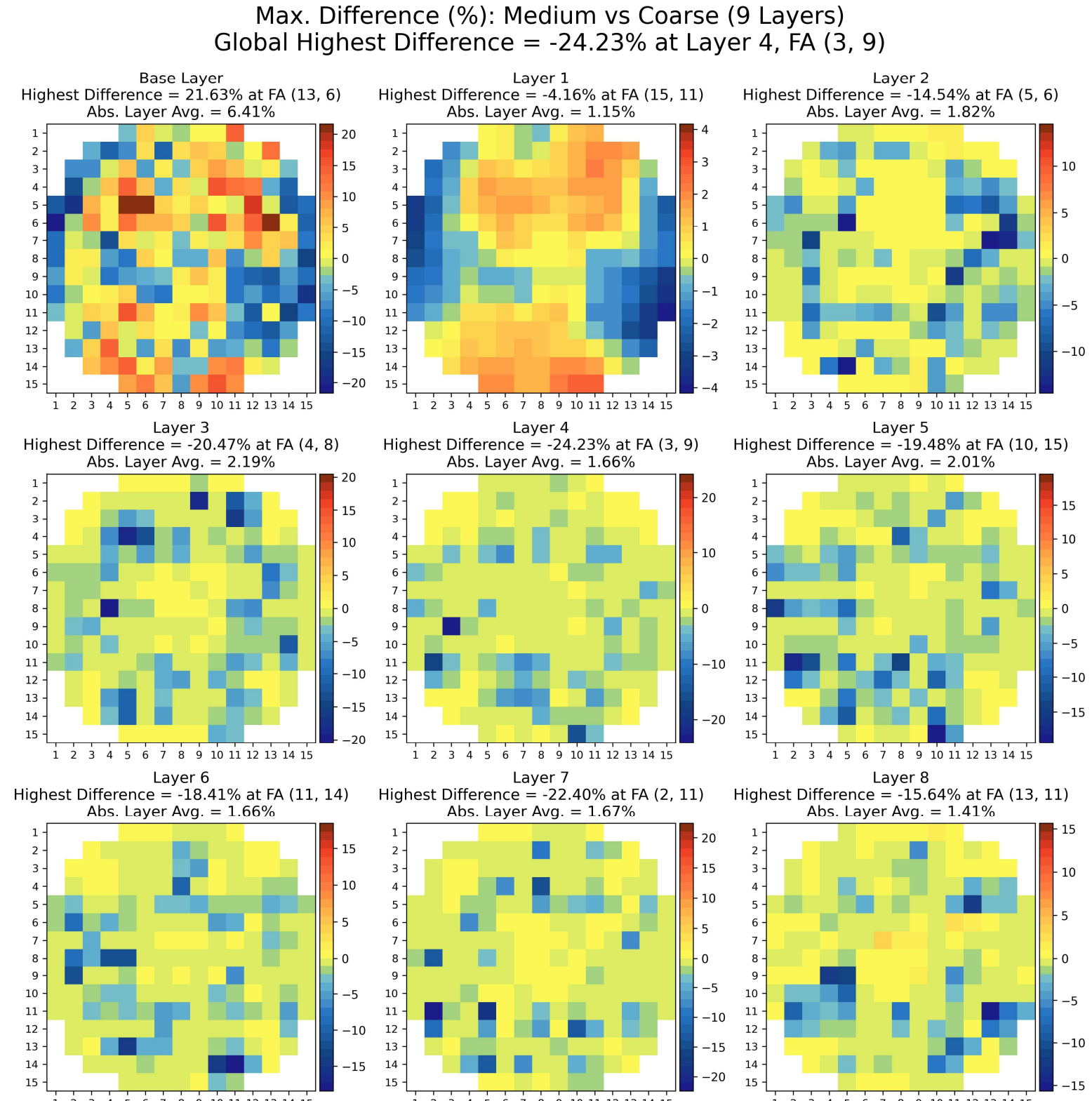

Figure A.15: Maximum error: Medium vs Coarse mesh comparison for all axial layers.

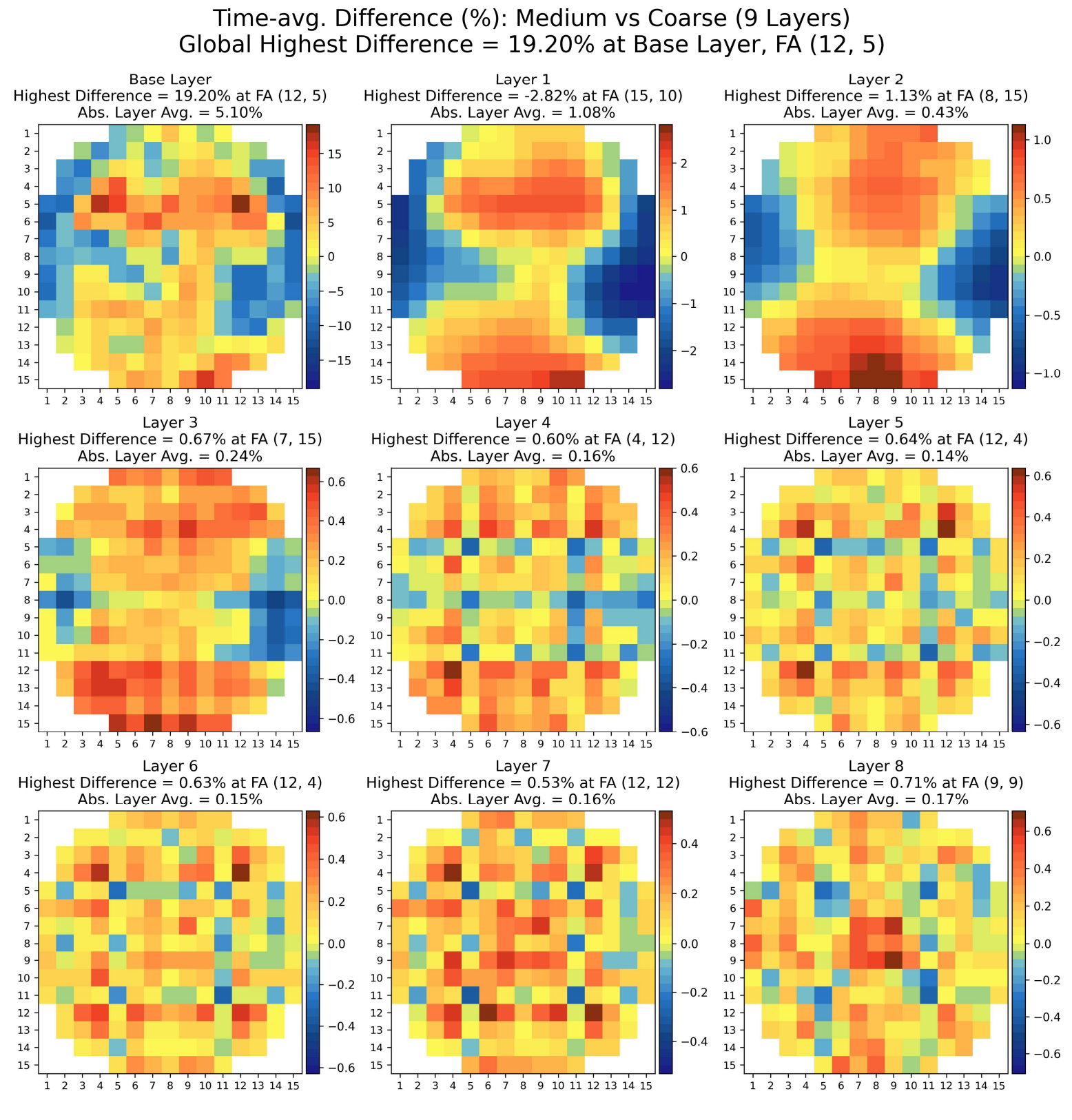

Figure A.16: Time-averaged error: Medium vs Coarse mesh comparison for all axial layers.